\documentclass[11pt]{book}
\usepackage{LaTex/iiit_thesis}
\usepackage{times}
\usepackage{epsfig}
\usepackage{amsmath}
\usepackage{amssymb}
\usepackage{latexsym}
\usepackage{graphicx}
\usepackage{multirow}
\usepackage{amsthm}
\usepackage{LaTex/algorithm2e}
\DeclareMathOperator{\sgn}{sgn}
\DeclareMathOperator{\sat}{sat}

\newtheorem{assum}{Assumption}
\usepackage[noadjust]{cite}
\newtheorem{remark}{Remark}
\newtheorem{theorem}{Theorem}
\newtheorem{mydef}{Definition}
\usepackage{epstopdf}
\usepackage{soul}

\counterwithin{mydef}{chapter}
\counterwithin{remark}{chapter}
\counterwithin{assum}{chapter}
\counterwithin{theorem}{chapter}

\graphicspath{{images/}{./}} 

\long\def\symbolfootnote[#1]#2{\begingroup%
\def\thefootnote{\fnsymbol{footnote}}\footnote[#1]{#2}\endgroup}
\renewcommand{\baselinestretch}{1.2}
\onecolumn

\begin{document}
\pagenumbering{roman}

\thispagestyle{empty}
\begin{center}
\vspace*{1.5cm}
{\Large \bf ADAPTIVE CONTROLLERS FOR QUADROTORS CARRYING UNKNOWN PAYLOADS}

\vspace*{2.75cm}
{\large Thesis submitted in partial fulfillment\\}
{\large  of the requirements for the degree of \\}

\vspace*{1cm}
{\it {\large Master of Science} \\
{\large in\\}
{\large Electronics and Communication Engineering \\}
{\large by Research \\}}

\vspace*{1cm}
{\large by}

\vspace*{5mm}
{\large VISWA NARAYANAN S.\\}
{\large 2019702013\\
{\small \tt VISWA.NARAYANAN@RESEARCH.IIIT.AC.IN}}

\vspace*{4.0cm}
{\psfig{figure=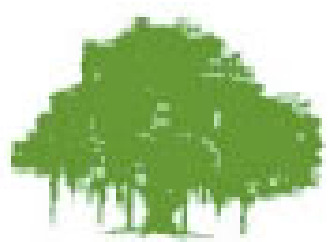,width=14mm}\\}
{\large International Institute of Information Technology\\}
{\large Hyderabad - 500 032, INDIA\\}
{\large AUGUST 2021\\}
\end{center}

\newpage
\thispagestyle{empty}
\renewcommand{\thesisdedication}{{\large Copyright \copyright~~VISWA NARAYANAN S., 2021\\}{\large All Rights Reserved\\}}
\thesisdedicationpage

\newpage
\thispagestyle{empty}
\vspace*{1.5cm}
\begin{center}
{\Large International Institute of Information Technology\\}
{\Large Hyderabad, India\\}
\vspace*{3cm}
{\Large \bf CERTIFICATE\\}
\vspace*{1cm}
\noindent
\end{center}
It is certified that the work contained in this thesis, titled ``Adaptive controllers for quadrotors carrying unknown payloads'' by VISWA NARAYANAN S., has been carried out under
my supervision and is not submitted elsewhere for a degree.

\vspace*{3cm}
\begin{tabular}{cc}
\underline{\makebox[1in]{}} & \hspace*{5cm} \underline{\makebox[2.5in]{}} \\
Date & \hspace*{5cm} Adviser: Prof. SPANDAN ROY
\end{tabular}
\oneandhalfspace

\newpage
\thispagestyle{empty}
\renewcommand{\thesisdedication}{\large To SRM Team Humanoid and the scientific community}
\thesisdedicationpage

\mastersthesis
\renewcommand{\baselinestretch}{1.5}

\chapter*{Acknowledgments}
\label{ch:ack}
Like Newton famously said, "If I have seen further, it is by standing on the shoulders of giants" I have come so far only with the guidance and support of all my well-wishers. I want to thank them and dedicate my thesis to each and every one of them.

First and foremost, I am grateful to Prof. K. Madhava Krishna for giving me an opportunity to explore and learn robotics at the Robotics Research Center, IIIT Hyderabad. Only because of that opportunity I had got a fundamental perspective on how to contribute to research. I am equally grateful to Dr. Spandan Roy for guiding me and supporting me to initiate my research and carry it on. Also, the love and support from my family in all the decisions that I have made are incomparable.

I am fortunate to have had great friends who have pushed me beyond my comfort zone and take research seriously. I am in debt to the trust that Hari Krishnan and Vikram Kumar have had in me while starting a new humanoid robotics team at SRM University. I belong to the community of roboticists only because the doors of SRM Team Humanoid opened for me. I am glad that I have met and been in contact with Aditya Sripada for all our technical and philosophical discussions.

I am thankful to Robotics Research Center (RRC), which welcomed me with open arms. The valuable mentoring by Gourav Kumar has given me a comprehensive overview of aerial robotics. Not to forget the significant partnership with Suraj in coursework and research that has helped me complete my degree in time. I thank everyone in RRC for the acceptance, inspiration and motivation you have given me from the beginning.

Finally, I have to thank the institute for providing a platform to aspiring minds to take up the dream of becoming a researcher and developing themselves in the process.

Thank you all.

\chapter*{Abstract}
\label{ch:abstract}
With the advent of intelligent transport, quadrotors are becoming an attractive solution while lifting or dropping payloads during emergency evacuations, construction works, etc. During such operations, dynamic variations in (possibly unknown) payload cause considerable changes in the system dynamics. However, a systematic control solution to tackle such varying dynamical behaviour is still missing. In this work, two control solutions are proposed to solve two specific problems in aerial transportation of payload, as mentioned below.

In the first work, we explore the tracking control problem for a six degrees-of-freedom quadrotor carrying different unknown payloads. Due to the state-dependent nature of the uncertainties caused by variation in the dynamics, the state-of-the-art adaptive control solutions would be ineffective against these uncertainties that can be completely unknown and possibly unbounded a priori. In addition, external disturbances such as wind while following a trajectory in all three positions and attitude angles. Hence, an adaptive control solution for quadrotors is proposed, which does not require any a priori knowledge of the parameters of quadrotor dynamics and external disturbances.  

The second work is focused on the interchanging dynamic behaviour of a quadrotor while loading and unloading different payloads during vertical operations. 
The control problem to maintain the desired altitude is formulated via a switched dynamical framework to capture the interchanging dynamics of the quadrotor during such vertical operations, and a robust adaptive control solution is proposed to tackle such dynamics when it is unknown.

The stability of the closed-loop system employing both the proposed controllers are studied analytically via Lyapunov theory. Further, the effectiveness of the proposed solutions is verified by deploying the controller on a realistic simulator under various scenarios and by testing the performance by comparing them with state-of-the-art controllers.


\tableofcontents
\listoffigures
\listoftables

\chapter{Introduction}
\label{ch:intro}
Robotics has been one of the significant fields of interest owing to science fiction and fantasies. The idea of machines replacing a human in everyday tasks is so beneficial because of two main reasons. Firstly, the inefficiency of humans in doing mundane tasks. A free-thinking human is inefficient in performing repetitive tasks due to constraints such as boredom, distraction etc. The other reason is that machines (or robots in specific) can transcend the mechanical limitations of a human. Humans somehow wanted to enslave machines for these two reasons, and hence these machines were named robots (meaning `forced labours'). 

Much similar to the evolution of living beings, there has been a notable evolution of robots too. Humans invented robots for specific operations when humanoid fantasies were far-fetched and impractical with the technologies of their time. Hence, robots came in numerous feasible mechanical forms with task-specific features very similar to various living things. One significant step of robotic evolution was the autonomy of a robotic system, which relieves humans from the loop and drastically reduces the need for manual labour.

The autonomy of robots (or autonomous systems in general) posed a need for a new science stream called control. A robotic system can be defined as an integrated unit of machinery that coordinates to perform a task. In this context, an autonomous system is a robot that can perform tasks without being manually controlled by a human.

Autonomous vehicles/ mobile robots are fantasies brought to life with the advancement in control techniques. Autonomous vehicles are extensively used not only for research purposes but also for commercial applications. Autonomous vehicles are of different forms ranging from cars to ships to quadrotors. While the use cases of aerial and marine robots are limited due to safety constraints, road vehicles (carrying humans) are equipped with an autopilot feature already. So, every major automobile manufacturer is shifting its focus towards autonomous vehicles.

Control of autonomous vehicles is a broad category ranging from a high-level control in choosing an optimal trajectory to the low-level control in deciding the appropriate actuator inputs. The problem statement of each level of control is different from one another. For example, the high-level control might be a Model Predictive Control, which would have to give desired setpoint velocities to actuators, assuming an ideal actuator model. However, when it comes to low-level control, the goal would be to calculate the desired forces and torques required to follow the velocity or position trajectory that has been given as an input to the controller while relying on prior information about the dynamics of the system. An even lower level controller would translate the force and torque inputs into current or fuel output.

\section{Aerial Robotics}
Aerial robots are one of the best examples to show how humans invented machines to surpass their physical limitations. Aerial transportation has allowed a species that perceives the universe in three dimensions to exploit the third dimension for efficient travelling. As modern-day researchers on autonomous vehicles rightfully say, ``the additional dimension would solve the exponentially increasing road traffic problems".

Aerial vehicles are broadly classified as fixed-wing and rotary-wing quadrotors based on their mechanical design. Unmanned aerial vehicles (UAV) is a breakthrough in the applications of aerial vehicles. UAVs are not necessarily autonomous. They can also be teleoperated. In both cases, control plays an incredible role. Similar to other applications, the control problem of UAVs is also a broad category. The focus of this particular research is on low-level control of quadrotors. Control of a UAV is exceptionally critical compared to that of rovers (ground robots) because there is no braking mechanism in aerial robots when the system goes unstable.

The mechanical advantage in the structure of the fixed-wing aircraft makes it convenient for the transport of cargo and humans. It is widely used for personal, commercial, research, and defence purposes. However, the major disadvantage of fixed-wing aircraft is its take-off mechanism, which needs a runway, non-holonomic design, and its inability to hover on a position in the air. These disadvantages make them infeasible in many applications such as indoor goods transfer, disaster relief, surveying etc. and make room for rotary-wing quadrotors.

\section{Dynamics of Quadrotors}
Modelling any robotic system can be divided into four sub-sections: model representation, reference frames, representation of state variables, and dynamic equations of the system. Though there are different methods to model the dynamics in each subsection, I would focus on the method used in my works.

\subsection{Model Representation}
For complete dynamic representation, including non-linearities, a set of differential equations is used. For simplified modelling to use a linear controller, state-space representation might be convenient. However, it would not be optimal to design nonlinear controllers. In general, the UAV dynamics can be of a time-invariant differential form, as shown in the following equation

\begin{align*}
    a_n\frac{d^n y}{dt^n} + a_{n-1}\frac{d^{n-1} y}{dt^{n-1}} + a_{n-2}\frac{d^{n-2} y}{dt^{n-2}} +...+ a_1\frac{dy}{dt} + a_0y = b u
\end{align*}
where the left hand side represents the state variable (the variable that needs to be controlled) $y$ multiplied by gains $a_i$ and its derivatives, and the right hand side represents the control variable $u$ scaled by a factor of $b$. Depending on which degree of control is required, the differential equation of sufficient order is chosen. For a robotic system, typically the control inputs are forces and torques to control position and velocity of each degree-of-freedom. Hence, a second order differential equation is chosen.

\subsection{Reference Frames}
A crucial step in modelling the dynamics is to identify the frames of reference. For quadrotor applications, two reference frames are considered for ease of representation: an inertial frame to represent the absolute position and orientation of the quadrotor and a body-fixed frame to represent the body forces and torques.

\subsubsection{Inertial Frame}
The inertial reference frame is where a body with zero net force acting on it is either at rest or in motion along a straight line. It is generally taken the ground fixed plane. So the frame is stationary if we neglect the movement of the Earth. It is also known as the World frame and is taken such that its X-axis is aligned with the forward direction, Z-axis in the upward direction and Y-axis is chosen based on right-hand rule. Differential equations to represent the system dynamics is written inertial frame reference.

\subsubsection{Body-Fixed Frame}
The body-fixed frame is fixed to the quadrotor with the same axes convention as in the inertial frame.  Every frame has three orthonormal axes passing through the center-of-mass (CoM). For convenience, these axes are chosen as the principal axes for the body-fixed frame similar to the inertial frame. Forces from the propellers act on the body-fixed frame. Considering the example of a quadrotor, the thrusts from the motors produce forces and torques in the body-fixed frame. So, the angular velocities are represented in the body-fixed frames. 

\begin{figure}
	\centering
	\includegraphics[width=0.5\textwidth]{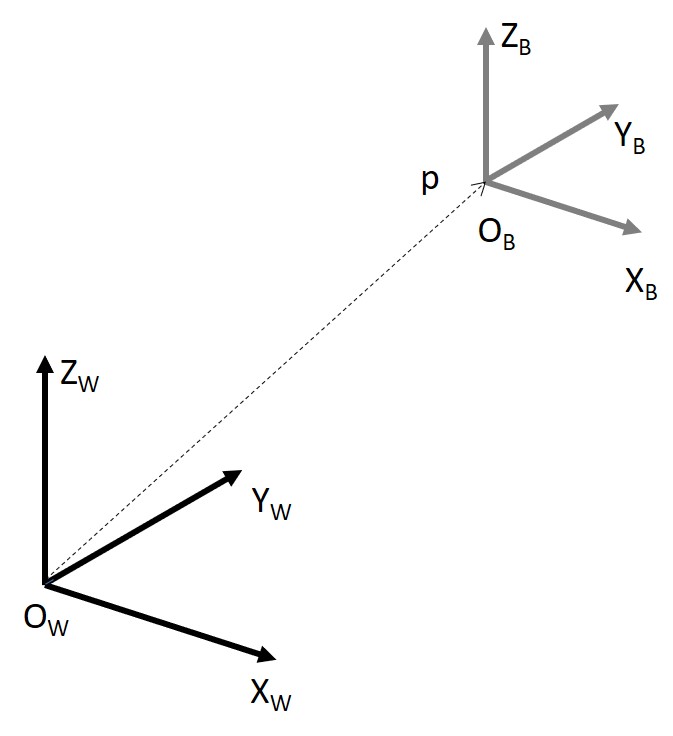}
	\caption{{A schematic of reference frames.}}\label{fig:frames}
\end{figure}

Figure \ref{fig:frames} shows the inertial and body-fixed frames with origins at $O_W$ and $O_B$, respectively. The position vector $p$ gives the relative position of CoM of the quadrotor (origin of body-fixed frame) in the inertial frame.

\subsection{Representation of State Variables}
Typically the pose of a quadrotor is represented by using the position vector $[x, y, z]^T$ and the orientation $[\phi, \theta, \psi]^T$, where $(\phi, \theta, \psi)$ represent roll, pitch and yaw angles respective. Though there are other conventions for representing the pose of a quadrotor, we would see a few methods used in the controller design.

\subsection{Quadrotor Dynamics}
The last part of the modelling is to derive the dynamic equations of the system.

The forces and torques in the quadrotor are only caused by the rotation of the propellers. Each propeller produces an upward force along the $Z_B$ axis in the body-frame and two different torques: one due to the upward force acting along $X_B$ and $Y_B$ axes, and the other is due to the spinning of each propeller, which causes a torque along the $Z_B$ axis. The quadrotor is designed symmetrically about the origin in the $X_BY_B$ plane to reduce complexity in dynamics. Also, the CoM, $O_B$ is maintained on the plane of the propeller.

\begin{figure}
	\centering
	\includegraphics[width=\textwidth]{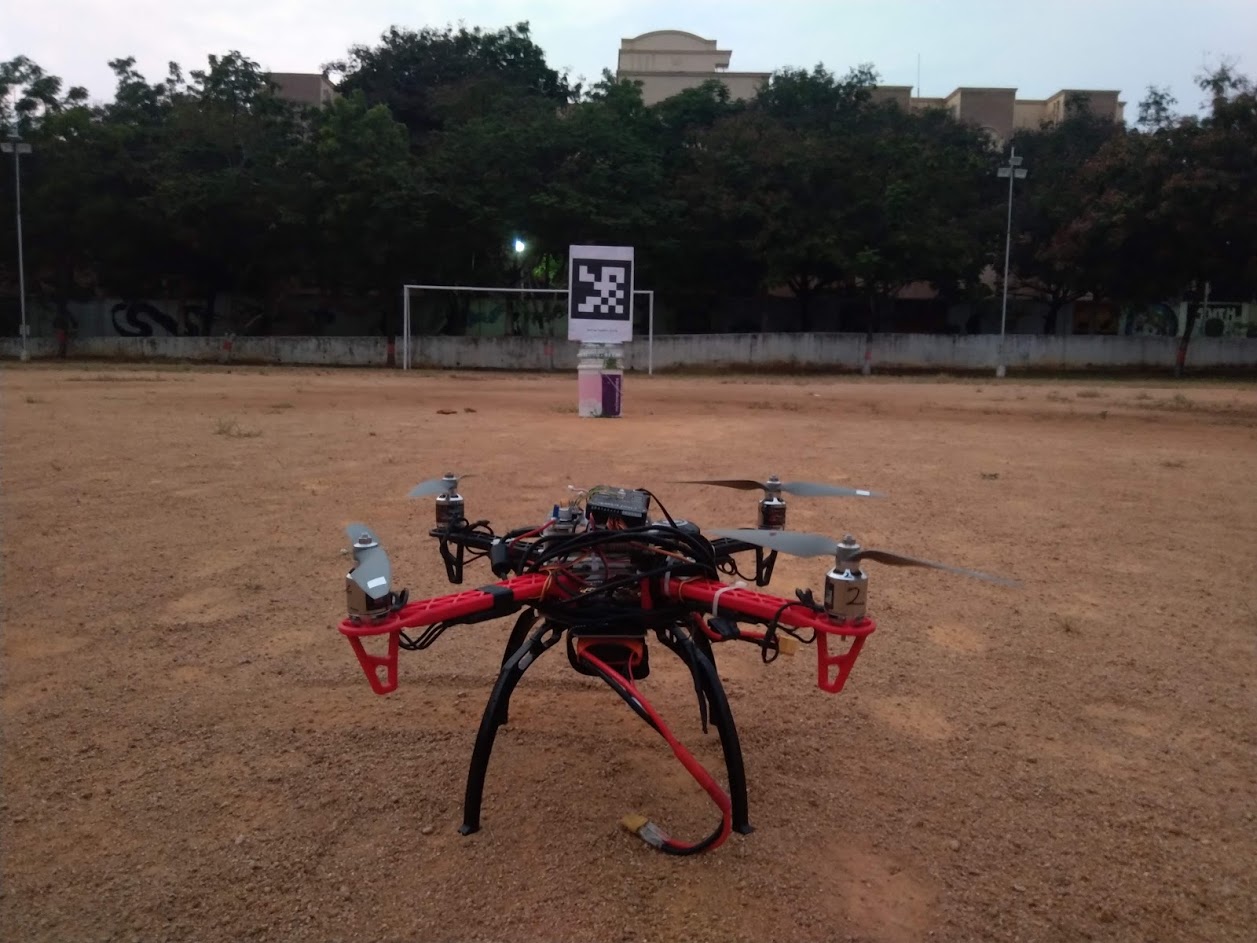}
	\caption{{A picture of Q450 quadrotor with ``x" configuration.}}\label{fig:q450_quad}
\end{figure}

The quadrotor's body frame is designed with its origin at the CoM of the quadrotor and $X_B$ and $Y_B$ axes along two adjacent arms (the connecting rod between the centre and the rotor), or 45 degrees on either side of an arm. The first configuration is called the "+" (plus) configuration, and the latter is called the "x" configuration (Fig. \ref{fig:q450_quad}). Here we demonstrate the dynamics of the quadrotor in the ``x" configuration.

When all propellers are spinning in the same direction with equal speed, the opposite propellers cancel out the torques along the $X_B$ and $Y_B$ directions. However, the torque along the $Z_B$-axis scales up, and the quadrotor starts to spin along the $Z_B$-axis while holding its position. So, the opposite propellers spin in the direction, while the adjacent ones spin in the opposite directions.

As the rotors are always facing downward, the quadrotor's thrust is always along the negative $Z_B$-axis. This essentially causes the quadrotor to move upwards against gravity when the body frame $Z_B$-axis is aligned with the world frame $Z_W$-axis. The velocity in $X_B$ and $Y_B$ directions are caused by pitching and rolling the quadrotor. To make the quadrotor pitch forward, the motors in the forward direction have to run slower than those in the backward direction. Similarly, to roll on the right side, the motors on the left side have to run faster than the ones on the left side and vice versa. Mismatch in the speeds of rotors can cause a difference between the torques in clockwise and anti-clockwise directions. This difference would make the quadrotor yaw. So, the sums of the squares of the speeds of the opposite motors are maintained to be equal.

\begin{figure}
	\centering
	\includegraphics[width=\textwidth]{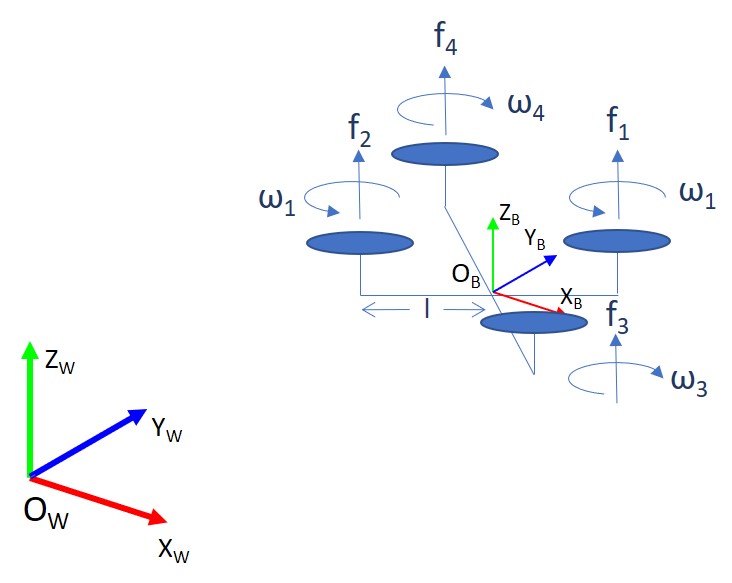}
	\caption{{A schematic representing a quadrotor's dynamics.}}\label{fig:quad_model}
\end{figure}

In this work we use the dynamic model of the quadrotor that is popularly used in works such as, \cite{lee2010geometric}, \cite{rotors_sim_2016} and \cite{mellinger2014traj}. The relationships between the thrust, $T_{th}$, torques $\tau_{\phi}, \tau_{\theta}, \tau_{\psi}$ and the rotor speeds $\omega_i$, where $i = 1,2,3,4$ are given by,

\begin{align}
    \begin{bmatrix}
        T_{th} \\ \tau_{\phi} \\ \tau_{\theta} \\ \tau_{\psi}
    \end{bmatrix} =
    \begin{bmatrix}
        C_T & C_T & C_T & C_T \\
        0 & lC_T & 0 & -lC_T \\
        -lC_T & 0 & lC_T & 0 \\
        -C_TC_M & C_TC_M & -C_TC_M & C_TC_M
    \end{bmatrix}
    \begin{bmatrix}
        \omega^2_1 \\
        \omega^2_2 \\
        \omega^2_3 \\
        \omega^2_4 
    \end{bmatrix}
\end{align}

where $C_T$ and $C_M$ are thrust constant and moment constant respectively, and $l$ is the arm-length (please refer to Fig \ref{fig:quad_model}).

When there are no external disturbances, these are the only forces and torques the quadrotor would experience other than gravity. As seen from the equations, there are only four control parameters that can be controlled, the angular velocities of the rotors. The control of these velocities would facilitate the control of upward thrust, roll, pitch and yaw. The linear motion in other directions can be controlled only using the control of roll and pitch. So, the quadrotor is an underactuated system with 4 control inputs for 6 DoF.

The linear forces acting on the system are gravitational force, $G$ along the negative $Z_W$-axis in the inertial frame and the thrust $T_{th}$, acting along the positive $Z_B$-axis in the body frame. Hence we use the linear dynamics of the system as given in \ref{pos_dyn}.

\begin{align}\label{pos_dyn}
    &m 
    \begin{bmatrix}
        \ddot{x}\\ 
        \ddot{y}\\ 
        \ddot{z}
    \end{bmatrix} + 
    \begin{bmatrix}
        0\\ 
        0\\ 
        mg
    \end{bmatrix} = \mathbf{R}^T 
    \begin{bmatrix}
        0\\ 
        0\\
        T_{th}
    \end{bmatrix}
\end{align}\\

where $g$ is the acceleration due to gravity, $m$ is the total mass of the quadrotor, $R$ is the rotational matrix that is used to map the forces from the inertial frame to the body fixed frame. As per properties of orthonormal matrices, $R^T$ maps vectors in body frame to the inertial frame. The rotational dynamics are given by the form,
\begin{align}
    &\frac{I_{xx}}{l}\ddot{\phi} + \frac{I_{zz}-I_{yy}}{l}\dot{\varphi}\dot{\psi} = \tau_\phi,\nonumber\\
&\frac{I_{yy}}{l}\ddot{\varphi} + \frac{I_{xx}-I_{zz}}{l}\dot{\phi}\dot{\psi}= \tau_\varphi, \nonumber\\
& I_{zz}\ddot{\psi} + (I_{yy}-I_{xx})\dot{\varphi}\dot{\phi}= \tau_\psi, 
\end{align} \label{att_sub}

where $l$ is arm-length of the rotor units; $I_{xx}, I_{yy},I_{zz}  $ are the inertia terms in $x,y$ and $z$ directions respectively. The rotation matrix $R$ is given by 
$$\mathbf{R} =\begin{bmatrix}
		c_\psi c_\varphi & s_\psi c_\varphi & -s_\varphi \\
		c_\psi s_\varphi s_\phi  - s_\psi c_\phi & s_\psi s_\varphi s_\phi  + c_\psi c_\phi & s_\phi  c_\varphi\\
		 c_\psi s_\varphi c_\phi + s_\psi s_\phi  &  s_\psi s_\varphi c_\phi - c_\psi s_\phi  & c_\varphi c_\phi
		\end{bmatrix}$$
where, $c_{(\cdot)},s_{(\cdot)}$ denote $\cos_{(\cdot)},\sin_{(\cdot)}$; $m$ is the mass of the overall system.

In this work, we have used the following quadrotor dynamics.

\begin{align} \label{el_pos}
   \mathbf{ M_p\ddot{p}} + \mathbf{G} &= \mathbf{\mathcal{T}_p}  \nonumber \\
    \mathbf{M_q(q)\ddot{q}} + \mathbf{C_q(\dot{q})\dot{q}} &= \mathbf{\mathcal{T}_q}
\end{align}

with

\begin{align}
    \mathbf{M_p} & = \begin{bmatrix}
        m & 0 & 0 \\
        0 & m & 0 \\
        0 & 0 & m
    \end{bmatrix}, ~
    \mathbf{M_q} = \begin{bmatrix}
        \frac{I_{xx}}{l} & 0 & 0 \\
        0 & \frac{I_{yy}}{l} & 0 \\
        0 & 0 & I_{zz}
    \end{bmatrix}, ~
    \mathbf{G} = \begin{bmatrix}
        0 \\
        0 \\
        mg
    \end{bmatrix}\nonumber \\
    \mathbf{C_q} &= \begin{bmatrix}
        0 & 0 & \frac{I_{zz} - I_{yy}}{l}\dot{\varphi} \\
        \frac{I_{xx} - I{zz}}{l}\dot{\psi} & 0 & 0 \\
        0 & \frac{I_{yy} - I{xx}}{l}\dot{\phi} & 0
    \end{bmatrix}, ~
    \mathbf{p} = \begin{bmatrix}
        x \\
        y \\
        z
    \end{bmatrix}, ~
    \mathbf{q} = \begin{bmatrix}
        \phi \\
        \varphi \\
        \psi
    \end{bmatrix}\nonumber \\
    \mathbf{\mathcal{T}_p} & = \begin{bmatrix}
        \tau_x \\
        \tau_y \\
        \tau_z
    \end{bmatrix}, ~
    \mathbf{\mathcal{T}_q} = \begin{bmatrix}
        \tau_{\phi} \\
        \tau_{\varphi} \\
        \tau_{\psi}
    \end{bmatrix}\nonumber \\
\end{align}

where, $\mathbf{M_p}$ and $\mathbf{M_q}$ are diagonal matrices that represent the mass and inertia of the quadrotor respectively; $\mathbf{C_q}$ is the Coriolis matrix containing the cross-coupling terms; $\mathbf{p}$ and $\mathbf{q}$ are the position and orientation vectors of the quadrotor in the inertial frame, $\mathbf{G}$ is the gravity vector and $\mathbf{\mathcal{T}_p}$ and $\mathbf{\mathcal{T}_q}$ are the force and moment vectors acting on the CoM of the quadrotor.


\section{Control of Quadrotors}
The initial works of quadrotor control were focused on reducing the complexity of the dynamics of the quadrotor. The simplest of all is the hovering mode control, where the quadrotor is assumed to have no disturbances across the $X_W$ and $Y_W$ axes. This type of control can be used to hover the quadrotor at any given height, and thus, only controlling the altitude of the quadrotor was sufficient. The earliest of control designs was well-proven Proportional-Integral-Derivative (PID) controller \cite{1302409,1389776, boua_pid}. 


Due to the limited applications of hovering mode, the next step in the development of quadrotor control was feedback linearization \cite{280180, smallanglemodel2, smallanglemodel3}: this approach is also known as small-angle variation control as it assumes that the variation in roll and pitch to be minimal.
This method is ideal for near hovering flying modes, and the adaptive controller is also designed for this specific condition \cite{lee2009asmc}. Though small-angle approximation based control is easy to implement, it might be unstable when the quadrotors need to perform aggressive manoeuvres.

 One of the significant pieces of research on the control of quadrotors is the geometric controller introduced by \cite{lee2010geometric}. This work proposed a partially decoupled dynamic model of the quadrotor (cf. Chapter 2 for details) and a dual-loop control system for the same. The geometric control in the manifold of rotation matrices makes it suitable for aggressive manoeuvres. Another notable work in this line came from \cite{mellinger2011minimum} and their group thereafter. The completely decoupled dynamic model of quadrotors, however is still in the theoretical phase and needs additional mechanical systems to realize it in practice (cf. tiltrotor mechanism \cite{8688053, 8815013}. An exceptional survey on the existing control strategies for autonomous quadrotors can be found in \cite{kimsurvey}.

Nevertheless, the aforementioned works (and references therein) depends heavily on the accuracy on system modelling and dynamics. In reality, it is tedious and almost impossible to model the exact dynamics of the quadrotor due to factors such as nonlinearities, inaccurate parametric knowledge, imbalance in the mass distribution, external disturbances etc. When quadrotors are deployed for applications such as surveying, manipulation, or payload transport \cite{aerial_mani, tang2015mixed, yang2019energy}, their dynamics keep changing due to payload variations. In this thesis, we particularly look into two types of aerial transportation and the control problems and solutions for them. We discuss these two scenarios below:

\subsection{Adaptive Path Tracking of Quadrotors for Aerial Transportation of Unknown Payloads} \label{motive_1}
Over the past two decades, quadrotors have been a source of considerable research interest owing to their advantages such as simple structure, vertical taking off and landing, rapid manoeuvring etc. \cite{fusini2018nonlinear, kapoutsis2013autonomous, nazaruddin2018communication}. Such advantages are crucial in various military and civil applications such as surveillance, fire fighting, environmental monitoring, to name a few \cite{invernizzi2019dynamic, invernizzi2019integral}. Most recently, global research is more and more interested in smart transport systems \cite{tang2015mixed, yang2019energy}.

One of the most common ways to attach a payload to a quadrotor is via a suspended cable, which allows the quadrotor to retain most of its agility \cite{tang2015mixed, sreenath2013trajectory, yang2019energy}. However, in indoor scenarios, especially in disaster sites, the quadrotor might need to manoeuvre through constrained altitudes, where using a suspended payload may not be optimal/desirable. In such scenarios, rigidly attaching a payload would be a preferred mode, which also provides the flexibility to autonomously pick up and drop the payload. From a research point of view, crucial control challenges arise when the mass and inertia of the quadrotor vary due to unknown payload, imprecise knowledge of the quadrotor parameters and unknown external disturbances: these uncertainties make the control design for a payload-carrying quadrotor challenging. 

\subsection{Dynamic Payload Lifting Operations: In-flight Payload Variation}\label{motive_2}
\begin{figure}[!h]
	\centering
	\includegraphics[width=\textwidth]{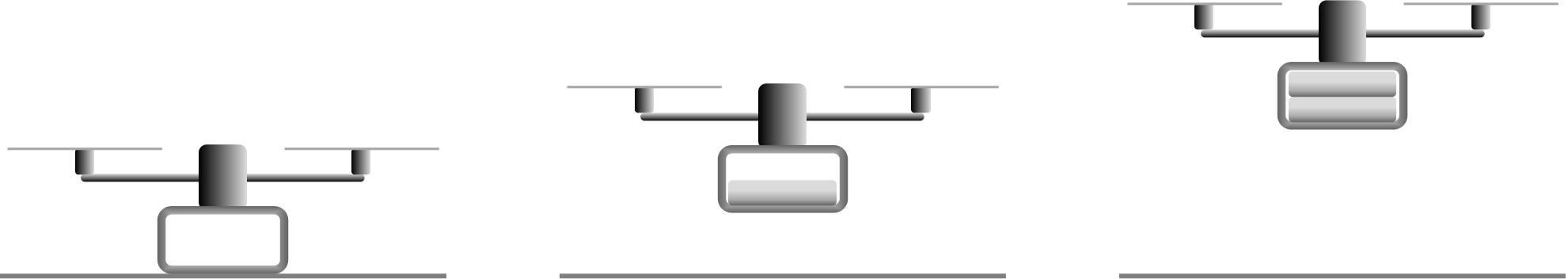}
	\caption{{A schematic representing a quadrotor ascending with dynamic payload.}}\label{fig:mot} 
\end{figure}
In the previous scenario, the payload mass does not vary during the flight. However, in a typical construction assignment or emergency evacuation, a quadrotor might be required to lift or drop varying payloads (construction and disaster relief materials), emergency evacuation from a high rise building, where humans are rescued from various floors. Dynamic variations in payload are orchestrated from these situations. One such lifting operation is sketched in Fig. \ref{fig:mot} via three phases: (i) in the first phase, the quadrotor rests on the ground without any payload; (ii) in the second phase, it starts to ascend with an initial payload; (iii) and in the third phase, a new payload is attached to the quadrotor as it ascends to its desired height. Similar phases can be observed during a construction scenario. In a reverse scenario to Fig. \ref{fig:mot}, a quadrotor may descend from a height while releasing payloads at different heights (e.g., dropping/throwing fire extinguishing materials at different floors). Clearly, the overall mass/inertia of the quadrotor changes (switches) during the interchange of these phases. 

Though robust controllers \cite{zhao2015nonlinear, nersesov2014estimation, sankaranarayanan2009control, xu2008sliding, Ref:17} might solve the problem of variation in dynamics to an extent, they also suffer when bounds of uncertainties are unknown. Hence, an adaptive controller is typically employed to solve these issues \cite{nicol2011robust, bialy2013lyapunov, dydek2012adaptive, ha2014passivity, tran2018adaptive, tian2019adaptive, zhao2014nonlinear, yang2019energy, roy2019adaptive, roy2020adaptive}. However, adaptive control typically requires structural knowledge of the system and cannot handle unmodelled dynamics, and such prerequisites are difficult to meet in the cases of unknown payload transport. Further, it is well-known that under a switched dynamics, typically arising during dynamic payload variation, conventional controllers are not suitable \cite{lai2018adaptive,yuan2018robust,yuan2017adaptive,lou2018immersion, chen2018global, ye2021robustifying}. In addition, to the best of our knowledge, no control solution exists to tackle the interchanging/switched dynamics of a quadrotor in an adaptive setting. This thesis aims to solve these control challenges.

\section{Preliminaries}
\subsection{Stability Notions}

An autonomous nonlinear system represented by the following dynamic equation
\begin{align*}
{\dot {x}}=f(x(t)),\;\;\;\;x(0)=x_{0}    
\end{align*}

where ${\displaystyle x(t)\in  \mathbb {R}^n}$ represents the system state vector is said to be stable, the function $f$ attains an equilibrium at the point $x = x_e$, i.e, $f(x_e) = 0$ and $ f'(x_e) = 0$.

The equilibrium point is said to be:
\begin{itemize}
    \item \textbf{Stable}, if there exists a bound $\epsilon > 0, \delta > 0$ such that $|x(0) - x_e| < \delta \implies |x(t) - x_e| < \epsilon, \quad \forall t \geq 0$. Stability makes sure the states starting from close enough to a bound $\delta$ will remain within the bound $\epsilon$.
    \item \textbf{Asymptotically stable}, if in addition to being stable, $\lim _{t\rightarrow \infty }\|x(t)-x_{e}\|=0$. Therefore, the system not only remains within the bound $\epsilon$ but also converges to the equilibrium point.
    \item \textbf{Exponentially stable}, if in addition with asymptotic stability there exists $ \alpha >0,\beta >0,\delta >0$ such that $\|x(t)-x_{e}\|\leq \alpha \|x(0)-x_{e}\|e^{-\beta t}, \quad \forall  t\geq 0$. Hence, system converges to the equilibrium at an exponential rate.
    \item \textbf{Uniformly Ultimately Bound (UUB):}
The solutions of $\dot{x} = f(x(t))$ is said to be uniformly ultimately bound with ultimate bound $b$ if $\exists b > 0,$ $c > 0$ and for every $0 < a < c,$ $\exists T = T(a,b) > 0$ such that
\begin{align}
    ||x(t_0)|| \leq a \implies  ||x(t)|| \leq b, \quad \forall t\geq t_o + T.
\end{align}
    
\end{itemize}

\section{Contributions of the Thesis}
Two different problems were highlighted earlier while a quadrotor is carrying a payload of unknown mass and inertia. While suspended payloads retain the dynamic properties of the quadrotor, they cannot be used in indoor scenarios and safety-critical operations. So, the following contributions are made in this thesis to solve the dynamic problems in quadrotors with rigidly attached payloads for their respective applications:
\begin{itemize}
    \item Based on the partly decoupled six degrees-of-freedom dynamic model, an adaptive sliding mode controller is proposed that tackles variation in quadrotor dynamics caused by uncertainties while carrying payloads of unknown mass and inertia, and external disturbances. The proposed controller does not require any a priori knowledge of the system dynamics and of uncertainty bound. The closed-loop system stability is analysed via the Lyapunov-based method. The performance of the proposed scheme is verified using a simulated quadrotor on the Gazebo framework and in comparison with the state-of-the-art controllers.
    \item In the previous case, the case of payload variation during the flight (e.g., payload drop during construction, disaster management scenario etc.) was not considered. Such in-flight payload variations give rise to the switched dynamics in the quadrotor system. A suitable switched adaptive controller, which does not require any a priori knowledge of the system dynamics and of uncertainties, is introduced for tracking any arbitrary altitude trajectory of the quadrotor. The closed-loop stability is analysed using multiple Lyapunov functions, and the effectiveness of the proposed controller is verified in simulation.
\end{itemize}

\section{Organization of the Thesis}
The thesis is organized into four chapters. A brief summary of each chapter is mentioned below.

\begin{itemize}
    \item \textbf{Chapter 1:} This introductory chapter gives an overview of aerial robotics, modelling the dynamics of a quadrotor and state-of-the-art control strategies. It briefly describes the motivation for this research, the problem orientation, the pertaining gaps in the literature, the main contributions and an outline of the thesis.
    \item \textbf{Chapter 2:} This chapter introduces the adaptive sliding mode controller for a quadrotor to address the problems involved while carrying payloads of unknown masses. The quadrotor dynamics are partly decoupled into two loops, with position tracking as the outer loop and attitude tracking as the inner loop. The uncertainties in both the loops are modelled and tackled using the adaptive sliding mode controller strategies. The stability of the controller is proved using the Lyapunov stability method and verified using a simulated quadrotor. The results are compared with the state-of-the-art controllers to show the superiority in the performance.
    \item \textbf{Chapter 3:} This chapter introduces the fundamentals of the switched mode controller for in-flight switched dynamics stemming from the addition and removal of unknown payloads. The quadrotor is modelled using hovering dynamics with control only on roll, pitch, yaw and altitude. An swtiched adaptive controller is designed to tackle the uncertainty. The closed-loop system is proved using dwell-time based multiple Lyapunov method and verified via simulation settings.
    \item \textbf{Chapter 4:} This chapter concludes the thesis by summarizing the various contributions brought out by this thesis. It also provides a brief discussion about the future work in this line of research.
    
\end{itemize}

\section{Symbols and Notations:}
The symbols and notations used in the following chapters are as shown in Table \ref{tab:notations}
\begin{table}[ht!]
    \centering
    \begin{tabular}{l l l}
        $\mathbb{R}$ & Real line  \\
        $\mathbb{R}^+$ & Real line of positive numbers \\
        $\mathbb{R}^n$ & Real space of dimension $n$ \\
        $\mathbb{R}^n$ & Real matrix of dimension $n \times n$ \\
        $\exists$ & there exists \\
        $\forall$ & for all \\
        $\mathbf{I}$ & Identity matrix \\ 
        $\mathbf{\Xi} > 0 (<0)$ & Positive (negative) definite matrix \\
        $\lambda_{\min}(\mathbf{\Xi})$  & Minimum eigen value of the matrix $\mathbf{\Xi}$ \\ 
        $\lambda_{\max}(\mathbf{\Xi})$  & Maximum eigen value of the matrix $\mathbf{\Xi}$ \\
        $||\mathbf{\Xi}||$ & Euclidean norm of the matrix $\mathbf{\Xi}$ \\
        $\sgn(x)$ & Signum of $x$ = $x/||x||$\\
        $\sat(x, \varpi)$ & Saturation of $x$ = 
        $\begin{cases}
            x/||x||, ~ if ~ ||x|| \geq \varpi \\
            x/ \varpi, ~ otherwise \\
        \end{cases}$
        
    \end{tabular}
    \caption{Nomenclature of various symbols and notations used in the thesis.}
    \label{tab:notations}
\end{table}


\chapter{Adaptive Path Tracking of Quadrotors for Aerial Transportation of Unknown Payloads}
\label{ch:asmc}
\section{Introduction}
In the quest to operate a quadrotor under various sources of uncertainties, the control community has inevitably looked into robust control \cite{xu2008sliding, sanchez2012continuous, derafa2012super, madani2007sliding} and adaptive control methods \cite{nicol2011robust, bialy2013lyapunov, dydek2012adaptive, ha2014passivity, mofid2018adaptive, tran2018adaptive, tian2019adaptive, zhao2014nonlinear, yang2019energy}. However, robust control methods \cite{xu2008sliding, sanchez2012continuous, derafa2012super, madani2007sliding} rely on a priori knowledge of bounds on uncertainties; on the other hand, the adaptive control designs \cite{nicol2011robust, bialy2013lyapunov, dydek2012adaptive, ha2014passivity, tran2018adaptive, tian2019adaptive, zhao2014nonlinear, yang2019energy} require a priori knowledge of system structures. Such constraints are often difficult to be satisfied in practice due to unknown parametric variations and external disturbances. 

To avoid a priori knowledge of bounds on uncertainties, researchers have recently applied adaptive sliding mode designs \cite{mofid2018adaptive, shtessel2012novel, utkin2013adaptive, obeid2018barrier} considering uncertainties to be a priori bounded. Nevertheless, when such assumption is not satisfied (e.g. state-dependent uncertainty), instability cannot be ruled out (cf. \cite{roy2020adaptive, roy2020towards}): in the attitude dynamics of a quadrotor, state-dependent uncertainty naturally occurs owing to the Coriolis terms.

 In view of the above discussions, an adaptive control solution for the quadrotor is still missing in the presence of \textit{completely unknown state-dependent uncertainty and external disturbances}.
Toward this direction, the proposed adaptive solution has the following major contributions:

\begin{itemize}
    \item An adaptive controller for quadrotor is formulated, which does not require any a priori knowledge of the system dynamic paramters, payload and of disturbances. 
    \item Differently from \cite{nicol2011robust, bialy2013lyapunov, dydek2012adaptive, ha2014passivity, mofid2018adaptive}, the control framework considers six degrees-of-freedom (DoF) dynamics with unknown external perturbations affecting both actuated and non-actuated sub-dynamics.
    \item The closed-loop stability of the system is analysed via Lyapunov-based method and comparative simulation results suggest significant improvement in tracking accuracy of the proposed scheme compared to the state of the art.
\end{itemize}

The rest of the chapter is organised as follows: Section \ref{sec:asmc_dyn} describes the dynamics of quadrotor; Section \ref{sec:asmc_cont} details the proposed control framework, while corresponding stability analysis is provided in Section \ref{sec:asmc_stab}; comparative simulation results are provided in Section \ref{sec:asmc_sim}.

\section{Quadrotor System Dynamics and Problem Formulation}\label{sec:asmc_dyn}
\begin{figure}[!h]
    \includegraphics[width=2.0 in, height=2in]{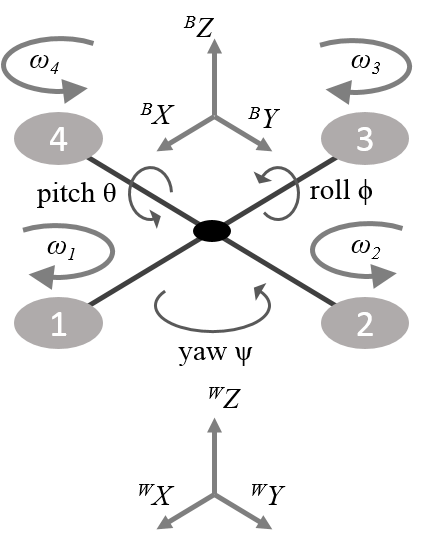}
    \centering
    \caption{Schematic of quadrotor with coordinate frames}
    \label{fig:quad_axis}
\end{figure}
	Let us consider the Euler-Lagrange system dynamics of a quadrotor model (cf. Fig. \ref{fig:quad_axis}), widely used in literature (\cite{bialy2013lyapunov})
	\begin{align}	
		m \ddot{p} + G + d_p &= \tau_p, \label{p_tau} \\
		J (q)\ddot{q} + C_q(q,\dot{q})\dot{q} + d_q &= \tau_q, \label{q_tau} \\
		\tau_p &= R_B^WU, \label{tau_conv}
	\end{align}
	where (\ref{p_tau}) and (\ref{q_tau}) are the position and attitude dynamics of the quadrotor, respectively. Various symbols in the dynamics (\ref{p_tau}) and (\ref{q_tau}) are described as follows: $m \in \mathbb{R}^{+}$ and $J(q) \in \mathbb{R}^{3 \times 3}$ represent the mass and inertia matrix, respectively; $p(t)\triangleq
	\begin{bmatrix}
	x(t) & y(t) & z(t)
	\end{bmatrix}^T \in\mathbb{R}^3$ is the position vector of the center of mass of the quadrotor; $q(t)\triangleq \begin{bmatrix}
	\phi(t) & \theta(t) & \psi(t) 
	\end{bmatrix}^T\in \mathbb{R}^3$ is the orientation/attitude (roll, pitch, yaw angles respectively); $G\triangleq
	\begin{bmatrix}
	0 & 0 & mg
	\end{bmatrix}^T\in \mathbb{R}^3$ is the gravitational force; $C_q(q, \dot{q}) \in \mathbb{R}^{3 \times3}$ represents the Coriolis matrix; $d_p(t)$ and $d_q(t)$ denote the unknown disturbances in position and attitude dynamics, respectively; $\tau_q \triangleq 
	\begin{bmatrix}
	u_2(t) & u_3(t) & u_4(t)
	\end{bmatrix}^T\in\mathbb{R}^3$ denotes the control inputs for roll, pitch and yaw; $\tau_p(t) \in \mathbb{R}^3$ is the generalized control input for position tracking in Earth-fixed frame, with $U(t)\triangleq
	\begin{bmatrix}
	0 & 0 & u_1(t)
	\end{bmatrix}^T\in \mathbb{R}^3$ being the force vector in body-fixed frame and $R_B^W \in\mathbb{R}^{3\times3}$ being the $Z-Y-X$ Euler angle rotation matrix describing the rotation from the body-fixed coordinate frame to the Earth-fixed frame, given by
	\begin{align}
	R_B^W =
	\begin{bmatrix}
	c_{\psi}c_{\theta} & c_{\psi}s_{\theta}s_{\phi} - s_{\psi}c_{\phi} & c_{\psi}s_{\theta}c_{\phi} + s_{\psi}s_{\phi} \\
	s_{\psi}c_{\theta} & s_{\psi}s_{\theta}s_{\phi} + c_{\psi}c_{\phi} & s_{\psi}s_{\theta}c_{\phi} - c_{\psi}s_{\phi} \\
	-s_{\theta} & s_{\phi}c_{\theta} & c_{\theta}c_{\phi}
	\end{bmatrix}, \label{rot_matrix}
	\end{align}
	where $c_{(\cdot)} , s_{(\cdot)}$ and denote $\cos{(\cdot)} , \sin{(\cdot)}$ respectively. 
	
	As per the celebrated Euler-Lagrange dynamics, the attitude dynamics of quadrotor satisfy the following properties \cite{bialy2013lyapunov}:
	
\noindent\textbf{Property 1.} The inertia matrix $J (q)$ is uniformly positive definite $\forall q$ and there exist $\underline{j},\overline{j} \in{R}^{+}$ such that $ \underline{j}I \leq J(q) \leq \overline{j}I$. \\
\textbf{Property 2.} $\exists \overline{c}_q, \overline{d}_p, \overline{d}_q \in\mathbb{R}^{+}$ such that $||C_q(q,\dot{q})|| \leq \overline{c}_q||\dot{q}||$, $||d_q|| \leq \overline{d}_q$, $||d_p|| \leq \overline{d}_p$. \\ 
\textbf{Property 3.} The matrix $(\dot{J} - 2C_q)$ is skew symmetric, i.e., for any non-zero vector $r$, we have $r^T(\dot{J} - 2C_q)r = 0$.

The following assumption highlights the available knowledge of various system parameters for control design:
\begin{assum}
The system dynamics terms $m,J,C_q,d_p,d_q$ and their bounds $\overline{c}_q, \overline{d}_p, \overline{d}_q$ are unknown for control design.
\end{assum}
\begin{remark}[Knowledge of system dynamics]
Assumption 1 is indeed a design challenge: the objective is to formulate a control framework without any knowledge of system parametric uncertainties and external disturbances. Assumption 1 overcomes the need for either precise knowledge of mass/inertia matrix (cf. \cite{xu2008sliding, sanchez2012continuous, derafa2012super, madani2007sliding}), or a priori knowledge of system structures and of bounds on external disturbances (cf. \cite{nicol2011robust, bialy2013lyapunov, dydek2012adaptive, ha2014passivity, tran2018adaptive, tian2019adaptive, zhao2014nonlinear, yang2019energy}).
\end{remark}
\begin{remark}[Position and attitude tracking co-design]\label{rem_co}
Tracking control problem of quadrotors can be broadly classified under two categories: (i) reduced-order model based design (cf. \cite{nicol2011robust, dydek2012adaptive, ha2014passivity, mofid2018adaptive}) and (ii) position and attitude tracking co-design (cf. \cite{bialy2013lyapunov, mellinger2011minimum}). Control designs relying on the first category ignore the non-actuated $(x,y)$ dynamics and define the control problem as tracking of only the actuated degrees-of-freedom (DoF) i.e. of $z$ and $(\phi, \theta ,\psi)$. Such approach is generally termed as `collocated design' in the literature of underactuated systems (cf. \cite{roy2020towards}). The second category relies on co-designing position and attitude tracking controller for the six DoF dynamics (\ref{p_tau})-(\ref{q_tau}). 
\end{remark}

In this work, we shall follow the co-design approach as the reduced-order based approach is conservative for relying on a priori boundedness of $(x,y)$ dynamics. Therefore, we take the following standard assumption:
\begin{assum}[\cite{bialy2013lyapunov, mellinger2011minimum, tang2015mixed}]\label{assum_des}
Let $p_d (t) \triangleq
	\begin{bmatrix}
	x_d (t) & y_d (t) & z_d (t)
	\end{bmatrix}^T$ and $\psi _d(t) $ be the desired position and yaw trajectories to be tracked, which are designed to be sufficiently smooth and bounded. Further, it is assumed that the following inequality holds for all time: $ - \frac{\pi}{2} < (\phi_d, \theta_d) < \frac{\pi}{2}$ radians.
\end{assum}
\begin{remark}[Desired roll and pitch]
In position and attitude tracking co-design, the desired roll ($\phi_d$) and pitch $(\theta_d$) trajectories are derived based on the computed position control input $\tau_p$ and the desired yaw trajectory $\psi_d$ (cf. \cite{mellinger2011minimum}) and it is discussed later.
\end{remark}
\textbf{Control Problem:} Under Properties 1-3, to design an adaptive controller to track a desired trajectory (cf. Assumption 2) without any knowledge of system dynamics parameters (cf. Assumption 1).

The following section provides a solution to this control problem.
\section{Controller Co-design}\label{sec:asmc_cont}
     The position and attitude tracking co-design approach consists of simultaneous design of an outer loop controller for position dynamics (\ref{p_tau}) and of an inner loop controller for attitude dynamics (\ref{q_tau}). Following this approach, the proposed control solution is elaborated in the following subsections. 
     
     \subsection{Outer Loop Controller}
Let us define the position tracking error as $e_p \triangleq p - p_d$ and
a sliding variable as
\begin{align}
     s_p &= \dot{e}_p + \Phi_p e_p \label{p_s}
 \end{align}
where $\Phi_{p}$ is a positive definite gain matrix. Multiplying the time derivative of (\ref{p_s}) by $m$ and using (\ref{p_tau}) yields
\begin{align}
     m\dot{s}_p = m(\ddot{p}-\ddot{p}_d + \Phi_p \dot{e}_p) = \tau_p + \varphi_p - G \label{dot_s_p}
\end{align}
where $\varphi_p
\triangleq -(d_p + m\ddot{p}_d - m\Phi_p\dot{e}_p)$. Using Property 2, we have
\begin{align}
 ||\varphi_p|| &\leq \overline{d}_p + {m}(||\ddot{p}_d|| + ||\Phi_p||  ||\dot{e}_p||).
\end{align}
Further, let us define $\xi_p \triangleq
\begin{bmatrix}
    e_p^T & \dot{e}_p^T
\end{bmatrix}^T$. Then using the inequality $||\xi_p|| \geq ||\dot{e}_p||$ and boundedness of the desired trajectories, we have
\begin{align}
    ||\varphi_p|| &\leq K_{p0}^* + K_{p1}^*||\xi_p||  \label{p_up_bound_K}
\end{align}
where $K_{p0}^* \triangleq {\overline{d}_p + m||\ddot{p}_d||}$ , $K_{p1}^* \triangleq m||\Phi_p||$ are \textit{unknown} finite scalars. The outer loop control law is designed as
\begin{subequations}\label{ip_out}
\begin{align}
    \tau_p (t) &= -\Lambda_p s_p (t) - \rho_p (t) \sgn(s_p(t)) + \hat{m}(t) g_p \label{p_con_tau}\\
    \rho_p(t) &= \hat{K}_{p0}(t) + \hat{K}_{p1}||\xi_p|| \label{p_con_rho}
\end{align}
\end{subequations}
where $\Lambda_p$ is a positive definite user-defined gain matrix; $g_p =  
    \begin{bmatrix}
        0 & 0 & g
    \end{bmatrix}^T$ is the gravity vector; $\hat{K}_{pi}$ and $\hat{m}$ are the estimates of $K_{pi}^*$ and ${m}$, $i=0,1$ respectively and they are evaluated via the following adaptive laws
    \begin{subequations}\label{adap_out}
\begin{align}
    \dot{\hat{K}}_{pi}(t) &= ||s_p(t)|| ||\xi_p(t)||^i - \alpha_{pi} \hat{K}_{pi}(t), ~\hat{K}_{pi}(0) > 0  \label{p_adap_K} \\
    \dot{\hat{m}}(t) &= - s_p(t)^T g_p - \alpha_m \hat{m}(t), ~ \hat{m}(0) > 0\label{p_adap_m}
\end{align}
\end{subequations}
where $\alpha_{pi}, \alpha_m  \in \mathbb{R}^{+}$ are user-defined design scalars. Note that eventually $U$ is applied to the system by transforming $\tau_p$ via the relation (\ref{tau_conv}) ($R^W_B$ is invertible rotational matrix), giving non-zero input only in $z$ direction.

\subsection{Inner Loop Controller}
To realize the inner loop controller, it is important to generate the desired roll ($\phi_d$) and pitch ($\theta_d$) angles. This process involves defining an intermediate coordinate frame as the first step (cf. \cite{mellinger2011minimum}):
\begin{subequations}\label{int_co}
\begin{align}
    z_B &= \frac{\tau_p}{||\tau_p||} \\
    y_A &= \begin{bmatrix}
    -s_{\psi_d} & c_{\psi_d} & 0
\end{bmatrix}^T \\
    x_B &= \frac{y_A \times z_B}{||y_A \times z_B||} \\
    y_B &= z_B \times x_B
\end{align}
\end{subequations}
where $y_A$ is the $y$-axis of the intermediate coordinate frame $A$; $x_B$, $y_B$ and $z_B$ are the $x$-axis, $y$-axis and $z$-axis of the body fixed coordinate frame. Given the desired yaw angle $\psi_d (t)$ and based on the computed intermediate axes as in (\ref{int_co}), $\phi_d (t)$ and $\theta_d (t)$ can be determined using the desired body frame axes as described in \cite{mellinger2011minimum}. 

Further, to achieve the attitude tracking control objective, the error in orientation/attitude is defined as \cite{mellinger2011minimum}
\begin{align}
    e_q &= {((R_d)^T R_B^W - (R_B^W)^T R_d)}^{v} \label{q_err} \\
    \dot{e}_q & = \dot{q} - R_d^T R_B^W \dot{q}_d
\end{align}
where $(.)^v$ represents \textit{vee} map, which converts elements of $SO(3)$ to $\in{\mathbb{R}^3}$ \cite{mellinger2011minimum} and $R_d$ is the rotation matrix as in (\ref{rot_matrix}) evaluated at ($\phi_d, \theta_d, \psi_d$).

The sliding variable $s_q \in{\mathbb{R}^3}$ for inner loop control is defined as
\begin{align}
     s_q = \dot{e}_q + \Phi_q e_q \label{q_s}
\end{align}
where $\Phi_{q}$ is a positive definite gain matrix. Multiplying the derivative of (\ref{q_s}) by $J$ and using (\ref{q_tau}) yield
\begin{align}
     J\dot{s}_q = J(\ddot{q}-\ddot{q}_d + \Phi_q{\dot{e}_q}) = \tau_q - C_q s_q + \varphi_q \label{dot_s_q}
 \end{align}
 where $\varphi_q 
 \triangleq -(C_q \dot{q} + d_q + J\ddot{q}_d - J\Phi_q\dot{e}_q - C_q s_q)$ represents the overall uncertainties in attitude dynamics. Using (28) and Properties 1 and 2 we have
 \begin{align}
     ||\varphi_{q}|| &\leq \overline{c}_q||\dot{q}||^2 + \overline{d_{q}} + \overline{j}(||\ddot{q}_d|| + ||\Phi_{q}|| ||\dot{e}_{q}||)  + \overline{c}_q||\dot{q}||(||\dot{e}_{q}|| + ||\Phi_{q}||||q||). \label{q_phi_bound}
 \end{align}
 Further, let us define $\xi_q \triangleq
	\begin{bmatrix}
	e_q^T & \dot{e}_q^T
	\end{bmatrix}^T$. Then using the inequalities $||\xi_q|| \geq ||e_q||$, $||\xi_q|| \geq ||\dot{e}_q||$, boundedness of the desired trajectories, and substituting $\dot{q} = \dot{e}_q + \dot{q}_d$ in $(\ref{q_phi_bound})$ yield
\begin{align}
    ||\varphi_{q}|| &\leq K_{q0}^* + K_{q1}^*||\xi_{q}|| + K_{q2}^*||\xi_{q}||^2, \label{q_up_bound_K}
\end{align}
where $K_{q0}^* \triangleq \overline{c}_q||\dot{q}_d||^2 + \overline{d}_q + \overline{j}||\ddot{q}_d||$, $K_{q1}^* \triangleq \overline{c}_q||\dot{q}_d||(3 + ||\Phi_q||) + \overline{j}||\Phi_q||$, $K_{q2}^* \triangleq {\overline{c}_q||\dot{q}_d||(2 + ||\Phi_q||)}$ are \textit{unknown} finite scalars. Hence, a state-dependent upper bound occurs naturally to the attitude tracking system. So, the inner loop control law is designed as
\begin{subequations}\label{ip_inn}
\begin{align}
	    \tau_q (t) &= -\Lambda_q s_q (t) - \rho_q (t) \sgn(s_q(t)), \label{q_con_tau} \\
	    \rho_{q}(t) &= \hat{K}_{q0}(t) + \hat{K}_{q1}(t)||\xi_q (t)|| + \hat{K}_{q2}(t)||\xi_q(t)||^2, \label{q_con_rho}
\end{align}
\end{subequations}
where $\Lambda_q$ is a positive definite user-defined gain matrix and $\hat{K}_{qi}$, $i=0,1,2$ are the estimates of ${K}_{qi}^*$ adapted via the following law:
\begin{align}
    \dot{\hat{K}}_{qi}(t) = ||s_q(t)||||\xi_q||^i - \alpha_{qi} \hat{K}_{qi}(t),~\hat{K}_{qi} (0) > 0, \label{q_adap_K}
\end{align}
where $\alpha_{qi} \in\mathbb{R}^{+}$, $i=0,1,2$ are user-defined scalars.
\begin{remark}[On state-dependent uncertainty] The inequalities (\ref{p_up_bound_K}) and (\ref{q_up_bound_K}) reveal that the state-dependencies occur inherently in the upper bound structures of uncertainties via $\xi_p$ and $\xi_q$. Therefore, conventional adaptive sliding mode designs such as \cite{mofid2018adaptive, shtessel2012novel, utkin2013adaptive, obeid2018barrier} are not feasible as state-dependent uncertainty cannot be bounded a priori. On the other hand, the gains $\rho_p$ in (\ref{p_con_rho}) and $\rho_q$ in (\ref{q_con_rho}) are designed according to the state-dependent uncertainty structures (\ref{p_up_bound_K}) and (\ref{q_up_bound_K}) respectively.
\end{remark}
\begin{remark}[On co-design of outer and inner loop]
Note that the outer and inner loop designs are not fully decoupled process: rather, the inner loop design process relies on $\tau_p$ from the outer loop. Therefore, these loop designs are coupled and simultaneously affect the quadrotor system: this makes the proposed design a more realistic one compared to decoupled dynamics based approaches (cf. \cite{nicol2011robust, dydek2012adaptive, ha2014passivity, mofid2018adaptive}) or autopilot based designs where often linearized or decouple dynamics are considered (cf. \cite{fari2019addressing,yang2019software}).
\end{remark}
\section{Stability Analysis of The Proposed Controller}\label{sec:asmc_stab}
\begin{theorem}
Under Properties 1-3 and Assumptions 1-2, the trajectories of the closed-loop systems (\ref{dot_s_p}) and (\ref{dot_s_q}) using the control laws (\ref{ip_out}) and (\ref{ip_inn}), in conjunction with the adaptive laws (\ref{adap_out}) and (\ref{q_adap_K}) are Uniformly Ultimately Bounded (UUB).
\end{theorem}
\textit{Proof.}
	Note that the solutions of the adaptive gains $( \ref{p_adap_K} )$ and $(\ref{q_adap_K})$ are given by \cite{roy2020adaptive}
\begin{align}
\vspace{-0.8cm}
    \hat{K}_{pi}(t) &= \underbrace{\exp(-\alpha_{pi} t) \hat{K}_{pi} (0)}_{\geq 0} + \underbrace{\int_{0}^{t} \exp(-\alpha_{pi}(t-\vartheta)) (|| {s_p} (\vartheta)|| ||\xi_p (\vartheta)||^{i}) \mathrm{d}\vartheta}_{\geq 0} \nonumber \\
    \Rightarrow~\hat{K}_{pi} (t) &\geq 0,~ i=0,1 ~\forall t\geq 0 \label{p_low_bound} \\
    \hat{K}_{qi}(t) &= \underbrace{\exp(-\alpha_{qi} t) \hat{K}_{qi} (0)}_{\geq 0}  + \underbrace{\int_{0}^{t} \exp(-\alpha_{qi}(t-\vartheta)) (|| {s_q} (\vartheta)|| ||\xi_q (\vartheta)||^{i}) \mathrm{d}\vartheta}_{\geq 0} \nonumber\\
    \Rightarrow~\hat{K}_{qi} (t) &\geq 0,~ i=0,1,2 ~\forall t\geq 0. \label{q_low_bound}
\end{align}
Closed-loop stability is analysed using the following Lyapunov function
\begin{align}
 V &= V_p + V_q, \label{overall_lyap}\\
\text{where}~    V_p &= \frac{1}{2}s_p^T m s_p + \frac{1}{2} \sum \limits_{i=0}^{1} (\hat{K}_{pi} - K_{pi}^*)^2 + \frac{1}{2} (\hat{m} - m)^2 \label{p_lyap}\\
    V_q &= \frac{1}{2}s_q^T J s_q + \frac{1}{2} \sum \limits_{i=0}^{2} (\hat{K}_{qi} - K_{qi}^*)^2 .\label{q_lyap}    
\vspace{-0.5cm}
\end{align}
Using $(\ref{dot_s_p})$ and $(\ref{dot_s_q})$, the time derivatives of $(\ref{p_lyap})$ and $(\ref{q_lyap})$ yield
\begin{align}
    \dot{V}_p &= s_p^T m \dot{s}_p + \sum \limits_{i=0}^{1} (\hat{K}_{pi} - K_{pi}^*) \dot{\hat{K}}_{pi} + (\hat{m} - m)\dot{\hat{m}} \nonumber \\
    &= s_p^T (\tau_p + \varphi_p + G )  + \sum \limits_{i=0}^{1} (\hat{K}_{pi} - K_{pi}^*) \dot{\hat{K}}_{pi} + (\hat{m} - m)\dot{\hat{m}} \label{p_lyap_dot_2} \\
    \dot{V}_q &= s_q^T J \dot{s}_q + \frac{1}{2}s_q^T \dot{J} s_q + \sum \limits_{i=0}^{2} (\hat{K}_{qi} - K_{qi}^*) \dot{\hat{K}}_{qi} \nonumber\\
    &= s_q^T (\tau_q - C_q s_q + \varphi_q) + \frac{1}{2}s_q^T \dot{J} s_q   + \sum \limits_{i=0}^{2} (\hat{K}_{qi} - K_{qi}^*) \dot{\hat{K}}_{qi}. \label{q_lyap_dot_2}
\end{align}
Utilizing the upper bound structure given by (\ref{p_up_bound_K}) and Property 2, and the control laws (\ref{ip_out}), (\ref{ip_inn}), the terms $\dot{V}_p$ in (\ref{p_lyap_dot_2}) and $\dot{V}_q$ in (\ref{q_lyap_dot_2}) are simplified to
\begin{align}
    \dot{V}_p &= s_p^T (- \Lambda_p s_p - \rho_p \sgn(s_p) + \hat{m} g_p + \varphi_p - G )  + \sum \limits_{i=0}^{1} (\hat{K}_{pi} - K_{pi}^*) \dot{\hat{K}}_{pi} + (\hat{m} - m)\dot{\hat{m}} \label{p_lyap_dot_3} \\
    \dot{V}_q &= s_q^T (- \Lambda_p s_p - \rho_p \sgn(s_p) + \varphi_q) + \frac{1}{2}s_q^T (\dot{J} - 2C_q) s_q + \sum \limits_{i=0}^{2} (\hat{K}_{qi} - K_{qi}^*) \dot{\hat{K}}_{qi}. \label{q_lyap_dot_3}
\end{align}
Property 3 implies that $s_q^T (J-2C_q) s_q = 0$. Then utilizing the upper bound structure (\ref{p_up_bound_K}) and (\ref{q_up_bound_K}), and the facts that $G=mg_p$, $\rho_p \geq 0, \rho_q \geq 0$ from (\ref{p_low_bound}) and (\ref{q_low_bound}) we have
\begin{align}
    \dot{V}_p &= - s_p^T \Lambda_p s_p - \sum \limits_{i=0}^{1} (\hat{K}_{pi} - K_{pi}^*) (||\xi_p||^i ||s_p|| - \dot{\hat{K}}_{pi}) + (\hat{m} - m)(s_p^Tg_p + \dot{\hat{m}}) \label{p_lyap_dot_4} \\ 
    \dot{V}_q &= - s_q^T \Lambda_q s_q - \sum \limits_{i=0}^{2} (\hat{K}_{qi} - K_{qi}^*) (||\xi_q||^i ||s_q|| -\dot{\hat{K}}_{qi}). \label{q_lyap_dot_4}
\end{align}
Using (\ref{adap_out}) and (\ref{q_adap_K}), we have
\begin{align}
    (\hat{K}_{pi} - K_{pi}^*) \dot{\hat{K}}_{pi} &= ||s_p|| (\hat{K}_{pi} - K_{pi}^*) ||\xi_p||^i  + \alpha_{pi} \hat{K}_{pi} K_{pi}^* - \alpha_{pi} \hat{K}_{pi}^2 \label{Kp_dot_exp}\\
    (\hat{m} - m) \dot{\hat{m}} &= - (\hat{m} - m) s_p^Tg_p  + \alpha_{m} \hat{m} m - \alpha_{m} \hat{m}^2 \label{m_dot_exp}\\
    (\hat{K}_{qi} - K_{qi}^*) \dot{\hat{K}}_{qi} &= ||s_q|| (\hat{K}_{qi} - K_{qi}^*) ||\xi_q||^i + \alpha_{qi} \hat{K}_{qi} K_{qi}^* - \alpha_{qi} \hat{K}_{qi}^2 \label{Kq_dot_exp}
\end{align}
Substituting (\ref{Kp_dot_exp})-(\ref{Kq_dot_exp}) into (\ref{p_lyap_dot_4}) and (\ref{q_lyap_dot_4}) yield
\begin{align}
    \dot{V}_p & \leq - \frac{\lambda_{\min}(\Lambda_p)||s_p||^2}{3}  + \sum \limits_{i=0}^{1} (\alpha_{pi} \hat{K}_{pi} K_{pi}^* - \alpha_{pi} \hat{K}_{pi}^2) + \alpha_{m} \hat{m} m - \alpha_{m} \hat{m}^2 \nonumber \\
    & \leq - \frac{\lambda_{\min}(\Lambda_p)||s_p||^2}{3}  - \sum \limits_{i=0}^{1} \left(\frac{\alpha_{pi}(\hat{K}_{pi} -  K_{pi}^*)^2}{2} - \frac{\alpha_{pi} {K_{pi}^*}^2}{2} \right)  \nonumber \\
    & \quad - \left(\frac{ \alpha_{m} (\hat{m} - m)^2}{2} - \frac{\alpha_{m} {m}^2}{2} \right) \label{p_lyap_dot_5} \\
    \dot{V}_q & \leq - \frac{\lambda_{\min}(\Lambda_q)||s_q||^2}{3} + \sum \limits_{i=0}^{2} (\alpha_{qi} \hat{K}_{qi} K_{qi}^* - \alpha_{qi} \hat{K}_{qi}^2) \nonumber \\
    & \leq - \frac{\lambda_{\min}(\Lambda_q)||s_q||^2}{3}  - \sum \limits_{i=0}^{2} \left(\frac{\alpha_{qi}(\hat{K}_{qi} -  K_{qi}^*)^2}{2} - \frac{\alpha_{qi} {K_{qi}^*}^2}{2} \right). \label{q_lyap_dot_5}
\end{align}
Further the definition of Lyapunov function yields
\begin{align}
    V_p & \leq \frac{{m}}{2} ||s_p||^2 + \frac{1}{2} \sum \limits_{i=0}^{1} (\hat{K}_{pi} - K_{pi}^*)^2 + \frac{1}{2} (\hat{m} - m)^2 \label{p_lyap_upbound} \\
    V_q & \leq \frac{\overline{j}}{2} ||s_q||^2 + \frac{1}{2} \sum \limits_{i=0}^{2} (\hat{K}_{qi} - K_{qi}^*)^2. \label{q_lyap_upbound}
\end{align}
From $(\ref{p_lyap_upbound})$ and $(\ref{q_lyap_upbound})$, the conditions $(\ref{p_lyap_dot_5})$ and $(\ref{q_lyap_dot_5})$ can be further simplified to
\begin{align}
    \dot{V}_p & \leq -\varrho_p V_p + \frac{1}{2}\left( \sum \limits_{i=0}^{1} \alpha_{pi} {K^*_{pi}}^2 + \alpha_{m} {m}^2  \right) \label{p_lyap_dot_6} \\
    \dot{V}_q & \leq -\varrho_q V_q + \frac{1}{2}\left( \sum \limits_{i=0}^{2} \alpha_{qi} {K^*_{qi}}^2 \right) \label{q_lyap_dot_6}
\end{align}
where $\varrho_p \triangleq  \frac{\min_i\{ \lambda_{\min}(\Lambda_p)/3, \alpha_{pi}, \alpha_{m} \}}{\max \{ {m}/2, 1/2 \}}$, $\varrho_q \triangleq  \frac{\min_i \lbrace \lambda_{min}(\Lambda_q)/3,  \alpha_{qi} \rbrace }{\max \lbrace \overline{j}/2, 1/2 \rbrace} > 0$ can be designed via (\ref{p_con_tau}), (\ref{adap_out}), (\ref{q_con_tau}), and (\ref{q_adap_K}). From (\ref{p_lyap_dot_6}) and (\ref{q_lyap_dot_6}), the upper bound for the time derivative of the overall Lyapunov function $\dot{V}$ can be obtained as
\begin{align}
    \dot{V} & \leq - \varrho V + \frac{1}{2} \left( \sum \limits_{i=0}^{1} \alpha_{pi} {K^*_{pi}}^2  \right) + \frac{1}{2} \left( \alpha_{m} {m}^2 \right)  + \frac{1}{2}\left( \sum \limits_{i=0}^{2} \alpha_{qi} {K^*_{qi}}^2 \right) \label{overall_V_dot_1}
\end{align}
where $\varrho = \min \{ \varrho_p, \, \varrho_q\}$. Defining a scalar $\kappa$ such that $0 < \kappa < \varrho$, $(\ref{overall_V_dot_1})$ is simplified to
\begin{align}
    \dot{V} &= - \kappa V - (\varrho - \kappa) V  + \frac{1}{2} \left( \sum \limits_{i=0}^{1} \alpha_{pi} {K^*_{pi}}^2  +  \alpha_{m} {m}^2 + \sum \limits_{i=0}^{2} \alpha_{qi} {K^*_{qi}}^2 \right). \label{overall_V_dot_upbound}
\end{align}
Defining a scalar $\mathcal{ \bar B} \triangleq \frac{\left( \sum \limits_{i=0}^{1} \alpha_{pi} {K^*_{pi}}^2  +  \alpha_{m} {m}^2 + \sum \limits_{i=0}^{2} \alpha_{qi} {K^*_{qi}}^2 \right)}{2(\varrho - \kappa)}$, it can be seen that $\dot{V} (t) < - \kappa V (t)$ when $V (t) \geq \mathcal{ \bar B}$, so that
\begin{align}
    V & \leq max \{ V(0), \mathcal{ \bar B} \}, \forall t \geq 0,
\end{align}
and the closed-loop system remains UUB.

\begin{remark}
 Control laws (\ref{p_con_tau}) and (\ref{q_con_tau}) are discontinuous in nature, which may cause chattering. Therefore, as standard for sliding mode designs, the control laws can be made continuous by replacing the `signum' function by a `saturation' function defined as $\sat(s_p,\varpi_p)=s_p/||s_p||$ (resp. $s_p/\varpi_p$) if $|| s_p || \geq \varpi_p$ (resp. $|| s_p || <\varpi_p$) where $\varpi_p \in \mathbb{R}^{+}$ is a small scalar used to avoid chattering. Similarly, $\sgn(s_q)$ can be modified. Such modifications do not change the closed-loop stability result albeit some minor modifications in stability analysis, which can be carried out in the same line as in \cite{roy2019overcoming} and thus, repetition is avoided.
\end{remark}
\section{Simulation Results and Analysis}\label{sec:asmc_sim}
The performance of the proposed controller is tested on a Gazebo simulation platform using the RotorS Simulator framework \cite{rotors_sim_2016} for ROS with the Pelican quadrotor model (mass of the quadrotor without any payload is set to be $2$ kg). The results of the proposed adaptive sliding mode controller (ASMC) is compared with the geometric controller \cite{lee2010geometric} (referred to as PD control hereafter) available in the RotorS framework, and sliding mode controller (SMC) \cite{xu2008sliding} (note that existing ASMC designs such as \cite{mofid2018adaptive} cannot be applied as, aside considering the reduced-order model, they are designed for a priori bounded uncertainty, which is not valid for a quadrotor).

The Gazebo model of the quadrotor can be actuated by commanding the angular velocities for the rotors. The angular velocities are calculated by solving the following relationship between thrust, moments and angular velocities \cite{mellinger2011minimum}:
    \begin{align}
        \begin{bmatrix}
            \omega_1\\ \omega_2 \\ \omega_3 \\ \omega_4
        \end{bmatrix} &= 
        K^{-1} 
        \begin{bmatrix}
            u_1 \\ u_2 \\ u_3 \\ u_4
        \end{bmatrix} \\
        K &= \begin{bmatrix}
            k_f & k_f & k_f & k_f\\ -l k_f & 0 & l k_f & 0 \\ 0 & -l k_f & 0 & l k_f \\ k_{\tau} & -k_{\tau} & k_{\tau} & -k_{\tau} 
        \end{bmatrix}
    \end{align}
    
where $k_f$ and $k_{\tau}$ are force and moment constants respectively of the motors; $l$ is the arm length of the quadrotor and $\omega_i$ is the angular velocity of $i^{th}$ motor.

To properly judge the performance of the proposed design, two different simulation scenarios are created in the following subsections. For both the scenarios, the control parameters of the proposed ASMC are selected to be:
$\Phi_p = diag \lbrace
    2.4, 2.4, 16 \rbrace$, $\Phi_q = diag \lbrace
    0.22 , 0.22 , 0.01 \rbrace$, $\Lambda_p = diag \lbrace
    1.2 , 1.2 , 0.8 \rbrace$, $\Lambda_q = diag \lbrace 
    4.5 , 4.5 , 3.5 \rbrace$, $\hat{K}_{p0}(0) = \hat{K}_{p1}(0) = 0.01$, $\hat{K}_{q0}(0) = \hat{K}_{q1}(0) = \hat{K}_{q2}(0) = 0.0001$, $\alpha_{p0} = \alpha_{p1} = 3$, $\alpha_{q0} = \alpha_{q1} = \alpha_{q2} = 50$, $\hat{m}(0) = 0.01$ and $\alpha_m = 0.1$, $\varpi_p=0.1$, $\varpi_q=1$. For a fair comparison, similar sliding surfaces are selected for SMC as in (\ref{p_s}) and (\ref{q_s}). The gains for geometric controller \cite{lee2010geometric} is selected to be same as in the RotorS package, which are optimized for the Pelican model. Initial position and attitude for the quadrotor are selected to be $x(0)=y(0)=0$, $z(0)=0.1$ and $\phi(0)=\theta(0)=\psi(0)=0$.

To perform the experiments, initial values of positions, angles and their derivatives are taken to be zero. The control parameters chosen for the geometric controller is their default parameters for Pelican drone, i.e., position gain $k_p = 4$, linear velocity gain $k_d = 2.7$; for pitch and roll, angular gain and angular rate gain are chosen $1$ and  $0.22$  respectively, and $0.035, 0.01$ are chosen as angular gain and angular rate gain for yaw, respectively. 

\subsection{Scenario 1: Aggressive Manoeuvre}
\begin{figure}[t!]
    \includegraphics[width=\textwidth]{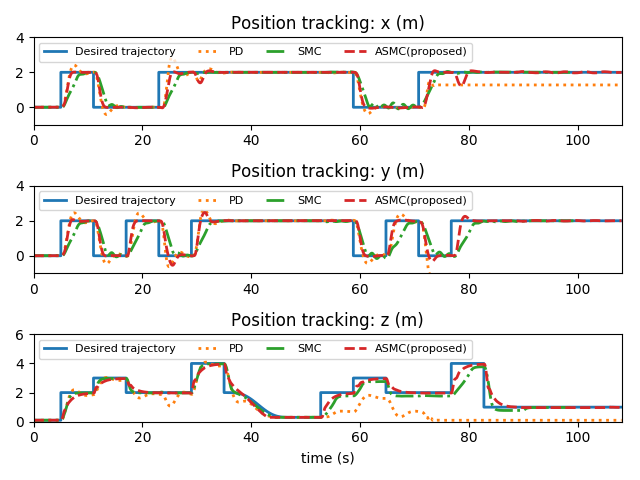}
    \centering
    \caption{Position tracking comparison for aggressive manoeuvre}
    \label{fig:traj_exp1}
\end{figure}

The objective of this scenario is to test the performance of various controllers when a quadrotor is to make aggressive manoeuvres. To this end, a set of way-points $(x_d, y_d, z_d)$ and $ \psi_d$ are commanded as input to the controllers (cf. Fig. \ref{fig:traj_exp1}). The list of waypoints provided is mentioned in the table \ref{tab:waypoints}. Once the last waypoint is reached, the quadrotor altitude is reduced to the height of the payload, and the payload is attached. Once again, the set of waypoints is given in the same time interval. These abrupt transitions between the positions demand aggressive manoeuvres for a fast response. For time duration $0\leq t<48$s, the set of waypoints are commanded without any payload attached to the quadrotor. Then, at $t=48$s, a payload of $0.5$ kg is added to the quadrotor, and the same set of waypoints are repeated.

\begin{table}[h!]
\renewcommand{\arraystretch}{1.1}
\caption{{List of waypoints for Scenario 1}}
\label{tab:waypoints}
		\centering
{
{	\begin{tabular}{c c c c c}
		\hline
		\hline
		
		 time (s) & $x$ (m) & $y$ (m) & $z$ (m)  & $\psi$ (deg) \\
		 \hline
		0.0 & 2.0 & 2.0  & 2.0 & 0.0 \\
		\hline
		6.0 & 0.0 & 0.0 & 3.0 & 0.0 \\
		\hline
		12.0 & 0.0 & 2.0 & 2.0 & 0.0 \\
		\hline
		18.0 & 2.0 & 0.0 & 2.0 & 90.0 \\
		\hline
		24.0 & 2.0 & 2.0 & 4.0 & 0.0 \\
		\hline
		\hline
\end{tabular}}}
\end{table}

\begin{figure}[t!]
    \includegraphics[width=\textwidth]{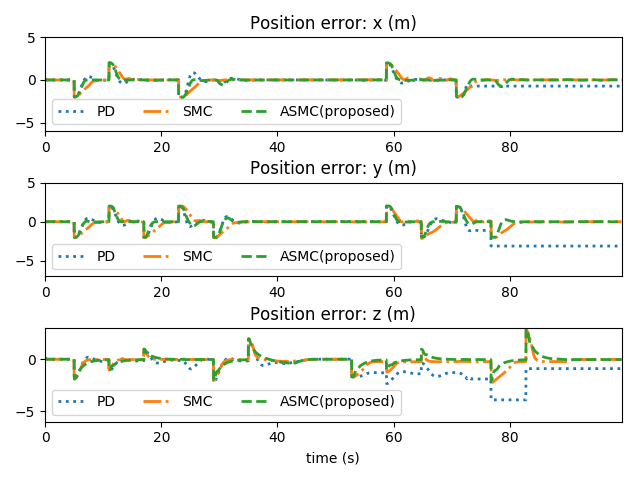}
    \centering
    \caption{Position tracking error comparison for aggressive manoeuvre}
    \label{fig:error_exp1}
\end{figure}

\begin{figure}[t!]
    \includegraphics[width=\textwidth]{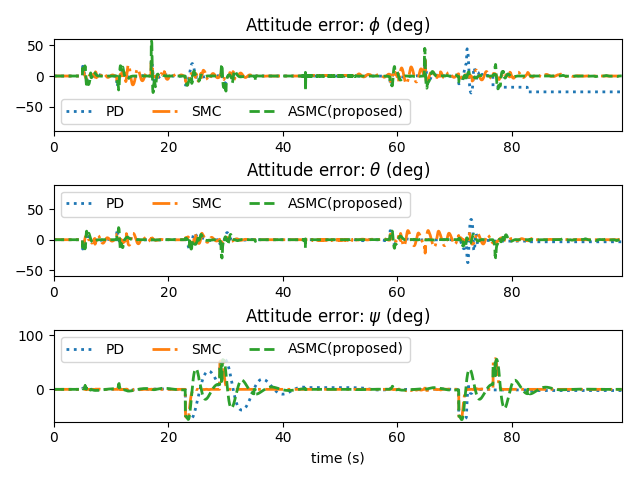}
    \centering
    \caption{Attitude tracking error comparison for aggressive manoeuvre}
    \label{fig:error_att_exp1}
\end{figure}

Figures \ref{fig:traj_exp1}-\ref{fig:error_att_exp1} depict the position tracking with various controllers and the corresponding errors incurred by them in position and attitude. It can be noted at $t\geq 72$s, the altitude ($z$ position) of the quadrotor using the geometric controller (PD) has become zero, i.e., the quadrotor has crashed on the floor during the manoeuvres after adding the payload. Reduction in the $z$ position error for the PD controller at $t>82$s should not be confused to be performance improvement: it is a result of reduction in the desired altitude $z_d$ (cf. Fig. \ref{fig:traj_exp1}), whereas the quadrotor still remains crashed on the ground. 

To demonstrate the performance of the controllers while carrying a payload, a performance comparison via root-mean-squared (RMS) error between SMC and ASMC is provided in Table \ref{table 1} after the payload is attached, i.e., for $t \geq 48$s (since the quadrotor crashed during the scenario for the geometric controller, its RMS error data is not tabulated). It clearly demonstrates the superior position tracking performance of the proposed scheme, where it has delivered a minimum performance improvement of $14.5 \%$. Despite position tracking is of more importance for transportation, it is also worth mentioning that attitude tracking of both the controllers is, however, almost similar. This is caused by the higher transients for ASMC compared to SMC during the sharp changes in the desired trajectory. This happens because the gains of ASMC need to re-adapt itself every time with the sharp changes in trajectories, while gains for SMC are fixed and less sensitive to abrupt changes. These transients are inherent for any adaptive design, but it helps them to adjust with unknown changes, while fixed-gain designs can fail when uncertainties lie beyond the a priori knowledge: such a situation is studied in the next scenario.  



Table shows the Root Mean Squared Error in the states $(x,y,z)$ and the overall position for the controllers. It can be observed that the proposed controller performs better than the other controllers.

\begin{table}[h!]
\renewcommand{\arraystretch}{1.1}
\caption{{Performance comparison for Scenario 1}}
\label{table 1}
		\centering
{
{	\begin{tabular}{c c c c c c c}
		\hline
		\hline
		& \multicolumn{3}{c}{RMS error (m)} & \multicolumn{3}{c}{RMS error (degree)}  \\ \cline{1-7}
		 position & $x$ & $y$  & $z$  & $\phi$ & $\theta$  & $\psi$  \\
		 \hline
		SMC & 0.47& 0.69  & 0.65 & 3.70 & 4.06  & 9.90 \\
		\hline
		ASMC (proposed)& {0.40} & {0.48}  & {0.42} & 3.80 & 2.67  & 10.06 \\
		\hline
		\hline
\end{tabular}}}
\end{table}
\clearpage

\subsection{Scenario 2: Dynamic payload}
\begin{figure*}
    \includegraphics[width=\textwidth]{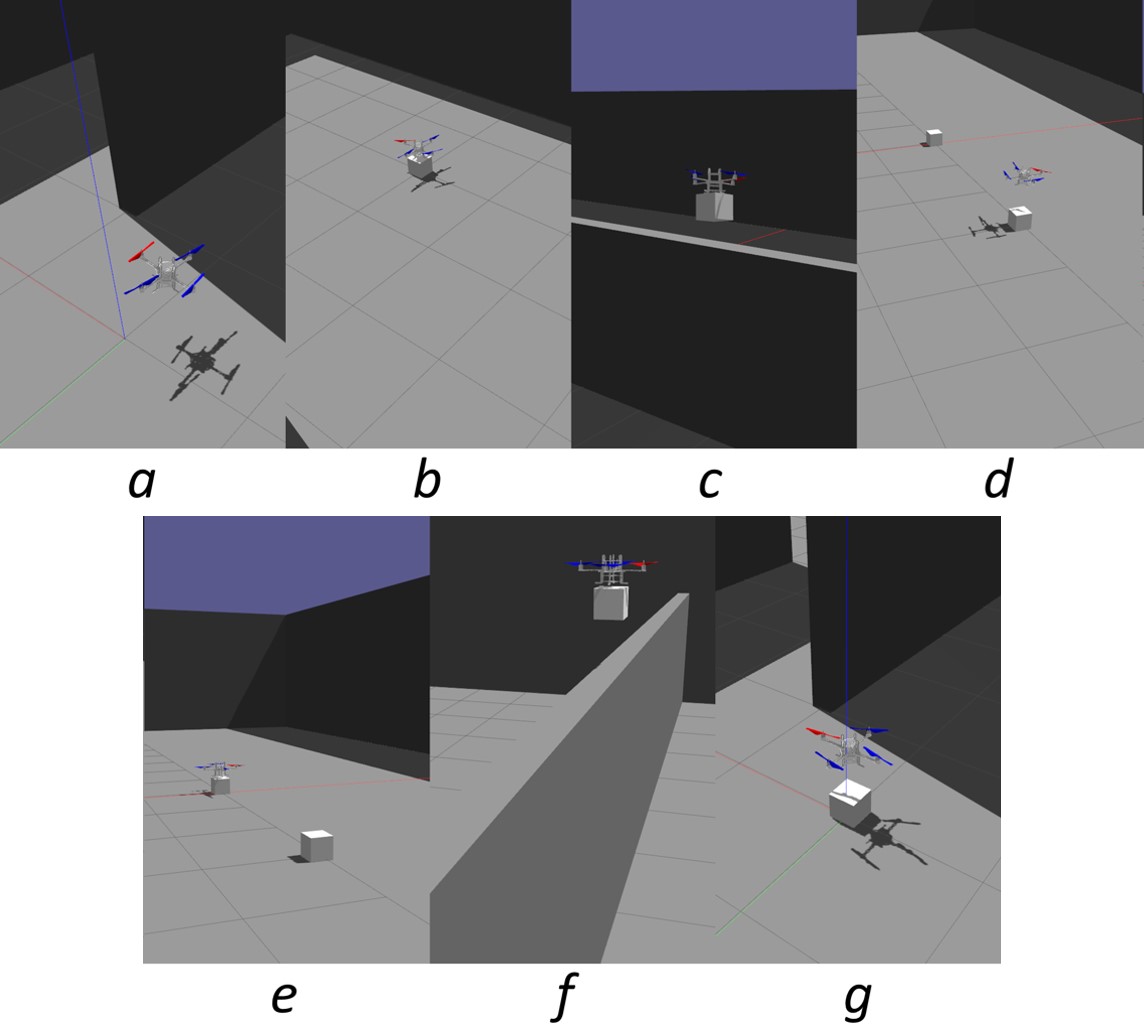}
    \centering
    \caption{Snapshots from experimental scenario 2 (with ASMC): quadrotor (a) starting from initial position (b) picking up first payload ($0.4$ kg) (c) moving over a wall with first payload (d) dropping the first payload (e) picking up second payload ($1$ kg) (f) moving over the wall with second payload (g) dropping the second payload at initial position.}
    \label{fig:scn_exp2}
\end{figure*}

\begin{figure}
    \includegraphics[width=\textwidth]{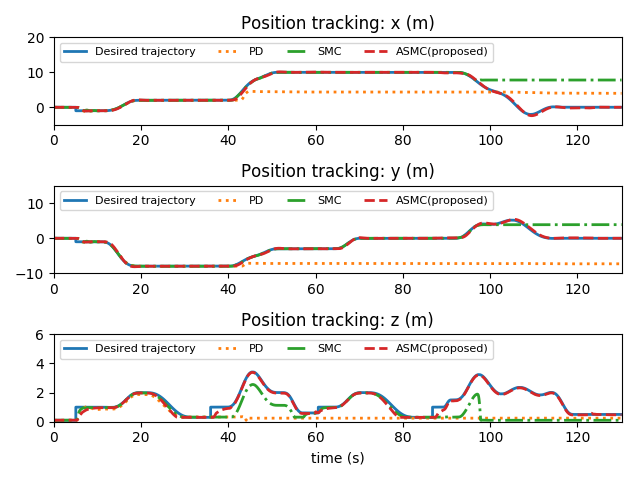}
    \centering
    \caption{Position tracking by various controllers with variable payloads}
    \label{fig:traj_exp2}
\end{figure}

\begin{figure}
    \includegraphics[width=\textwidth]{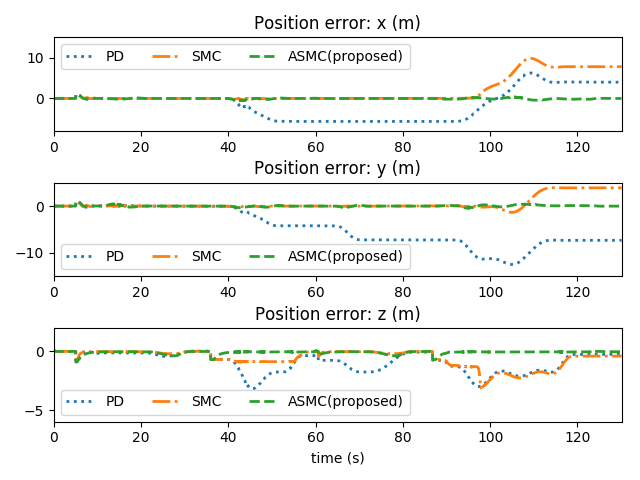}
    \centering
    \caption{Position Tracking error comparison with variable payloads}
    \label{fig:error_exp2}
\end{figure}

In a real-life aerial transport scenario, payloads can vary according to application requirements (e.g., human evacuations in an emergency). Accordingly, a second experimental scenario is created in the `indoor environment' available in the RotorS simulator, wherein performance of the various controllers under dynamic payload variation is tested while following a minimum snap trajectory \cite{mellinger2011minimum}. In this scenario, the quadrotor is required to perform the following actions sequentially (cf. Fig. \ref{fig:scn_exp2}):
\begin{itemize}
    \item From an initial position the quadrotor (unloaded) moves to the first payload location;
    \item it picks up a payload of $0.4$ kg (at approx $t=32$s) and an altitude of 1.0 m is given as the setpoint;
    \item it moves over a wall to the other side;
    \item it delivers the payload (at approx $t=58$s) and holds its altitude of 0.6 m;
    \item then, it moves towards a second payload of $1.0$ kg;
    \item it picks up the payload (at approx $t=83$s) and holds an altitude of 1.0 m;
    \item then, it moves over the wall back to the initial position;
    \item it delivers the payload (at approx $t=123$s) and holds its altitude of 0.5 m.
\end{itemize}
Moving over the walls with payload is an important phase for this experiment as it tests the capability of the controllers to maintain a designated altitude: such test cases are important for aerial transport in urban scenarios.
Masses of the first and second payload are chosen to be $0.4$ kg and $1.0$ kg. Figures \ref{fig:traj_exp2}, \ref{fig:error_exp2} show the states $(x, y, z)$ for different controllers with respect to time along with the commanded trajectory and their tracking errors respectively. The first payload is loaded at $t=32$s and unloaded at $t=58$s, while the second payload is loaded at $t=83$s and dropped at $t=123$s.

\begin{table}[!h]
\renewcommand{\arraystretch}{1.1}
\caption{{Performance of ASMC for Scenario 2}}
\label{table 2}
		\centering
{
{	\begin{tabular}{c c c c c c c}
		\hline
		\hline
		& \multicolumn{3}{c}{RMS error (m)} & \multicolumn{3}{c}{RMS error (degree)}  \\ \cline{1-7}
		 position & $x$ & $y$  & $z$  & $\phi$ & $\theta$  & $\psi$  \\
		\hline
	& {0.159} & {0.162}  & {0.159} & 1.278 & 1.072  & 0.311 \\
		\hline
		\hline
\end{tabular}}}
\end{table}

\begin{figure}
    \includegraphics[width=\textwidth]{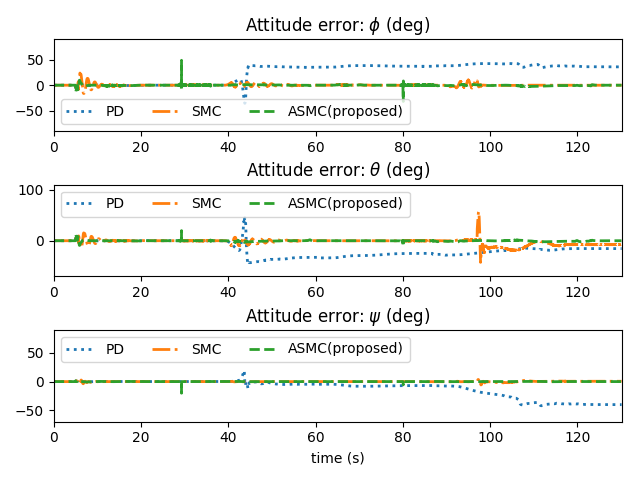}
    \centering
    \caption{Attitude tracking error comparison with variable payloads}
    \label{fig:error_att_exp2}
\end{figure}

Figures \ref{fig:traj_exp2}-\ref{fig:error_att_exp2} show the position responses with various controllers and their corresponding tracking errors. The position responses in the geometric controller become constant for $t>45$s, the quadrotor topples while moving with the first payload. A similar incident can be observed with SMC for $t> 92 $s: though the quadrotor could manage to lift the first payload, it could not lift the second payload high enough to cross the wall. This happened because SMC relies on a priori fixed gains based on predefined uncertainty bounds; when the payload becomes heavier, the controller cannot adjust its gain to adapt with higher uncertainty. In such a scenario, thanks to the adaptive gains as in (\ref{adap_out}) and (\ref{q_adap_K}), the proposed ASMC can tackle uncertainties without any a priori knowledge and thus, outperforms other controllers. The performance of ASMC is compiled in Table \ref{table 2}. The desired trajectories in this scenario being a smoother one (cf. \cite{mellinger2011minimum}) compared to the one in Scenario 1, transients in the attitude response for ASMC are visibly absent except at the instants when payloads are added. Similar to Scenario 1, since the quadrotor crashed during the scenario for SMC and PD, their performances are not tabulated.




\chapter{Dynamic Payload Lifting Operation}
\label{ch:switched_mode}
\section{Introduction}

In the previous chapter, it was considered that the payloads do not change during the flight. However, in many practical applications, payload mass varies abruptly: for example, dropping relief material, adding construction material during vertical movement (cf. Fig. \ref{fig:dyn}). Such abrupt changes give rise to switched dynamics: switched systems are an important class of hybrid systems that consists of subsystems with continuous dynamics, where at any given time instant, one subsystem remains active while others being inactive. Switching between active and inactive subsystems is governed by a switching law \cite{liberzon2003switching}.

In the quest to operate a quadrotor under various sources of parametric uncertainty, the control regime, after initial model based designs (cf. the survey paper \cite{rubisurvey} for the evolutions of various designs), has inevitably moved to some notable adaptive control works \cite{nicol2011robust, bialy2013lyapunov, dydek2012adaptive, ha2014passivity, mofid2018adaptive, tran2018adaptive, tian2019adaptive, zhao2014nonlinear, yang2019energy, roy2021artificial}. However, no adaptive design addresses the issue of switched dynamics as discussed earlier. On the other hand, extending these works to a switched dynamics scenario would rely on finding a common Lyapunov function, which again is difficult (if at all possible) to achieve in practical systems \cite{lai2018adaptive}. 

Therefore, a relevant question arises whether the existing switched adaptive control designs \cite{lai2018adaptive,yuan2018robust,yuan2017adaptive,lou2018immersion, chen2018global,8362915, 7782779, roy2019reduced, roy2019simultaneous, roy2020vanishing} (and references therein) can be applied to a switched quadrotor system while carrying an uncertain (and possibly unknown) payload. Unfortunately, existing literature does not present a positive answer to this: though \cite{roy2019reduced, roy2019simultaneous} do not rely on the structure of system dynamics in contrast to state of the art, they still require bound knowledge of the uncertain mass/inertia terms to design either control law \cite{roy2019simultaneous} or switching law \cite{roy2019reduced}. 

In light of the above discussions, an adaptive switched framework that can tackle completely unknown dynamics of quadrotor in a switched dynamics setting is still missing. Toward this direction, the proposed switched adaptive solution has the following major contributions:
\begin{itemize}
	\item A switched dynamics for quadrotor is formulated which suitably represents the changing dynamics during a vertical (lifting/dropping) operation under dynamic variation of payload.
	\item Compared to the state-of-the-art, the proposed adaptive law and switching law do not require any a priori knowledge of the system dynamics parameters and payload. Further, differently from \cite{nicol2011robust, mofid2018adaptive}, the effects of cross-coupling terms arising from the lateral motion are considered as state-dependent uncertainty.

\end{itemize} 

The rest of the chapter is organized as follows: Section 3.2 describes the switched quadrotor systems and highlights various issues in the state of the art; Section 3.3 details the proposed control framework, while the corresponding stability analysis is carried out in Section 3.4 and a simulation study is provided in Section 3.5.
\section{System Dynamics and Problem Formulation}
\begin{figure}
	\centering
	\includegraphics[width=\textwidth]{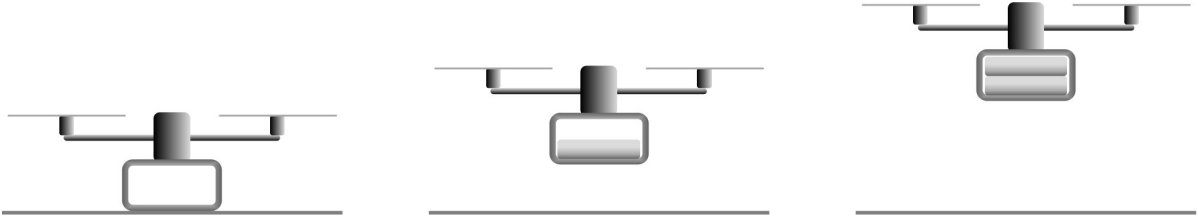}
	\caption{{Schematic depicting vertical operation with dynamic payload.}}\label{fig:dyn} 
\end{figure}

Let us consider the dynamics of quadrotor as given in Ch. 1:

\begin{align}\label{sw_pos_dyn}
    &m 
    \begin{bmatrix}
        \ddot{x}\\ 
        \ddot{y}\\ 
        \ddot{z}
    \end{bmatrix} + 
    \begin{bmatrix}
        0\\ 
        0\\ 
        mg
    \end{bmatrix} = \mathbf{R}^T 
    \begin{bmatrix}
        0\\ 
        0\\
        T_{th}
    \end{bmatrix}
\end{align}
\begin{align}
    &\frac{I_{xx}}{l}\ddot{\phi} + \frac{I_{zz}-I_{yy}}{l}\dot{\varphi}\dot{\psi} = \tau_\phi,\nonumber\\
&\frac{I_{yy}}{l}\ddot{\varphi} + \frac{I_{xx}-I_{zz}}{l}\dot{\phi}\dot{\psi}= \tau_\varphi, \nonumber\\
& I_{zz}\ddot{\psi} + (I_{yy}-I_{xx})\dot{\varphi}\dot{\phi}= \tau_\psi, 
\end{align} \label{sw_att_sub}
with the rotation matrix $\mathbf R$ is given by 
$$\mathbf{R} =\begin{bmatrix}
		c_\psi c_\varphi & s_\psi c_\varphi & -s_\varphi \\
		c_\psi s_\varphi s_\phi  - s_\psi c_\phi & s_\psi s_\varphi s_\phi  + c_\psi c_\phi & s_\phi  c_\varphi\\
		 c_\psi s_\varphi c_\phi + s_\psi s_\phi  &  s_\psi s_\varphi c_\phi - c_\psi s_\phi  & c_\varphi c_\phi
		\end{bmatrix}$$
where, $c_{(\cdot)},s_{(\cdot)}$ denote $\cos_{(\cdot)},\sin_{(\cdot)}$; $m$ is the mass of the overall system.

Multiplying both side of the position subsystem (\ref{sw_pos_dyn}) with the rotation matrix, $\mathbf{R}$ yields
\begin{subequations}\label{pos_dyn_mod}
	\begin{align}
	&	m\ddot{x}c_{\psi}c_{\varphi} + m\ddot{y}s_{\psi}c_{\varphi} - m\ddot{z}s_{\varphi} - mgs_{\varphi} = 0, \\
	&	m\ddot{x}(c_{\psi}s_{\varphi}s_{\phi} - s_{\psi}c_{\phi}) + m\ddot{y}(s_{\psi}s_{\varphi}s_{\phi} + c_{\psi}c_{\phi}) + m\ddot{z}s_{\phi}c_{\varphi} + mgs_{\phi}c_{\varphi} = 0 , \\
	&	m\ddot{x}(c_{\psi}s_{\varphi}c_{\phi} + s_{\psi}s_{\phi}) + m\ddot{y}(s_{\psi}s_{\varphi}c_{\phi} - c_{\psi}s_{\phi}) + m\ddot{z}c_{\varphi}c_{\phi} + mgc_{\varphi}c_{\phi} = T_{th}.
	\end{align}
\end{subequations}
After such rearrangements of the dynamics, we are now ready to define the (non-switched) four degrees-of-freedom dynamics structure suitable for collocated design as
\begin{equation}\label{sys_non_sw}
\mathbf M( \mathbf q)\ddot{ \mathbf q}+\mathbf C(\mathbf q,\dot{\mathbf q})\dot{\mathbf q}+\mathbf G(\mathbf q)+\mathbf{H}( \mathbf q)\ddot{ \mathbf{q}}_{\mathbf u}=\boldsymbol \tau,
\end{equation} 
where
\begin{align}
&\mathbf M =
\begin{bmatrix}
m(c_{\varphi}c_{\phi}) & 0 & 0 & 0 \\
0 & \frac{I_{xx}}{l} & 0 & 0\\
0 & 0 & \frac{I_{yy}}{l} & 0\\
0 & 0 & 0 & I_{zz}\\	
\end{bmatrix},~
\mathbf G = 
\begin{bmatrix}
mg(c_{\varphi}c_{\phi}) \\ 0 \\ 0 \\ 0
\end{bmatrix} \\
& \mathbf C = 
\begin{bmatrix}
0 & 0 & 0 & 0 \\ 
0 & 0 & 0 & \frac{(I_{zz} - I_{yy})}{l}\dot{\varphi}\\
0 & \frac{(I_{xx} - I_{zz})}{l}\dot{\psi} & 0 & 0 \\ 
0 & 0 & {(I_{yy} - I_{xx})}\dot{\phi}& 0
\end{bmatrix}	,\\
& \mathbf{H}=
\begin{bmatrix}
m(c_{\psi}s_{\varphi}c_{\phi} + s_{\psi}s_{\phi})  & m(s_{\psi}s_{\varphi}c_{\phi} - c_{\psi} s_{\phi}) \\ 
0 & 0\\
0 & 0 \\ 
0 & 0 
\end{bmatrix},\\
& \boldsymbol \tau = 
\begin{bmatrix}
T_{th} \\ \tau_\phi \\ \tau_\varphi \\ \tau_\psi
\end{bmatrix}, ~\mathbf q = 
\begin{bmatrix}
z \\ \phi \\ \varphi \\ \psi
\end{bmatrix},~\mathbf{q_u} = 
\begin{bmatrix}
x \\ y
\end{bmatrix}.
\end{align}

During a vertical operation, a quadrotor needs to maintain a given (possibly time-varying) height ($z$) and attitude $(\phi, \varphi, \psi)$, at a fixed $(x,y)$ position. Under such scenario, and based on the underactuated dynamics model as in (\ref{sys_non_sw}), a switched quadrotor dynamics (cf. Fig. \ref{fig:dyn}) for vertical operations can be represented as 
\begin{equation}\label{sys_1}
\mathbf M_{\sigma}( \mathbf q)\ddot{ \mathbf q}+\mathbf C_{\sigma}(\mathbf q,\dot{\mathbf q})\dot{\mathbf q}+\mathbf G_{\sigma}(\mathbf q)+\mathbf{H}_\sigma( \mathbf q)\ddot{ \mathbf{q}}_{\mathbf u}+\mathbf{ d}_{\sigma}=\boldsymbol \tau_{\sigma},
\end{equation} 
where $\mathbf{q}=\lbrace z, \phi, \varphi, \psi \rbrace$ and $\mathbf{q_u}=\lbrace x,y \rbrace$; for each subsystem $\sigma$, $\mathbf{M_{\sigma}(q)}\in\mathbb{R}^{4\times 4}$ is the mass/inertia matrix; $\mathbf C_{\sigma}(\mathbf q,\dot{\mathbf q})\in\mathbb{R}^{4 \times 4}$ denote the Coriolis, centripetal terms; $\mathbf G_{\sigma}(\mathbf q)\in\mathbb{R}^{4}$ denotes the gravity vector; $\mathbf{H}_\sigma( \mathbf q)\ddot{ \mathbf{q}}_{\mathbf u}\in\mathbb{R}^{4}$ represents the vector of cross-coupling terms stemming from $(x,y)$ sub-dynamics; $\mathbf{ d_{\sigma}}(t) \in \mathbb{R}^4$ denotes bounded external disturbance and $\boldsymbol \tau_{\sigma} \in \mathbb{R}^4$ is the control input. The mapping from $\mathbf \tau$ to thrust and torques of individual motors are discussed in Sect. \ref{sec:asmc_sim}. 

Here $\sigma(t) : [0~\infty) \mapsto \Omega$ is a piecewise constant function of time, called the switching signal, taking values in $\Omega=\lbrace 1,2, \cdots,N \rbrace $. The following class of slowly-switching signals is considered: 
\begin{mydef}
	Average Dwell Time (ADT) \cite{hespanha1999stability}: For a switching signal $\sigma(t)$ and each $t_2 \geq t_1 \geq 0$, let $N_{\sigma}(t_1,t_2)$ denote the number of discontinuities in the interval $[t_1,t_2)$. Then $\sigma(t)$ has an average dwell time $\vartheta$ if for a given scalar $N_0 >0$
	\begin{align*}
	N_{\sigma}(t_1,t_2) \leq N_0 + (t_2-t_1)/\vartheta,~~ \forall t_2 \geq t_1 \geq 0
	\end{align*}
	where $N_0$ is termed as chatter bound.
\end{mydef}

As the quadrotor dynamics (\ref{sys_1}) follows the celebrated Euler-Lagrange dynamics formulation \cite{tang2015mixed}, each subsystem in (\ref{sys_1}) presents a few interesting properties (cf. \cite{spong2008robot}), which are later exploited for control design as well as stability analysis:

\noindent \textbf{Property 1:} $\exists \overline c_{\sigma }, \overline g_{\sigma }, \overline h_{\sigma },\overline{d}_{\sigma} \in \mathbb{R}^{+}$ such that $||\mathbf C_{\sigma}(\mathbf q,\dot{\mathbf q})|| \leq \overline c_{\sigma } ||\dot{\mathbf q}||$, $||\mathbf{G_{\sigma}(q)}|| \leq \overline g_{\sigma }$, $||\mathbf{H}_\sigma( \mathbf q)|| \leq \overline h_{\sigma }$ and $||\mathbf{d_{\sigma}}(t)|| \leq \overline{d}_{\sigma}$.\\
\textbf{Property 2:} The matrix $\mathbf{M_{\sigma}(q)}$ is symmetric and uniformly positive definite $\forall \mathbf{q}$, implying that $\exists \underline m_{\sigma}, \overline m_{\sigma} \in \mathbb{R}^{+}$ such that
\begin{equation}\label{prop 3}
0 < \underline m_{ \sigma} \mathbf I \leq \mathbf{M_{\sigma}(q)} \leq \overline m_{ \sigma} \mathbf I .
\end{equation}

A few remarks regarding system (\ref{sys_1}) follow, which aid in designing the proposed switching control. 
\begin{remark}[(Uncertainty)]
	As a design challenge, the switched system (\ref{sys_1}) is considered to be unknown in the sense that the knowledge of $\mathbf{M}_\sigma,\mathbf{C_\sigma,H_\sigma, G_\sigma, d}_{\sigma}$ and their corresponding bounds, i.e., $\underline{m}_{\sigma}, \overline{m}_{\sigma}, \overline c_{\sigma }, \overline g_{\sigma }, \overline h_{\sigma },\overline{d}_{\sigma}$ are \textit{completely unknown}. 
\end{remark}
\begin{remark}[(Collocated Design)]\label{remark_collocated}
	Note that a quadrotor dynamics is essentially underactuated, where $x$ and $y$ positions represent the non actuated coordinates. During a vertical operation (i.e., a near hovering condition), there exist time-scale separation between $(z, \phi, \varphi, \psi)$ and $(x,y)$ dynamics, where the former evolves faster than the later \cite{dydek2012adaptive}. Therefore, it is standard to track only the actuated coordinates $(z, \phi, \varphi, \psi )$ of a quadrotor during such operation \cite{mofid2018adaptive, nicol2011robust}. In the literature of underactuated systems, such control design, i.e., tracking of only actuated coordinates, is called \textit{collocated design} (cf. \cite{shkolnik2008high,spong1994partial}). 
\end{remark}
\begin{remark}[(Cross-coupling terms)]
	Notwithstanding slow evolution of the lateral (i.e., $(x,y)$) motion compared to the vertical ($z$) and attitude $( \phi, \varphi, \psi )$ motions due to time-scale separation, ignoring its effect is conservative: especially, variation in payload and perturbation in attitude may cause some lateral motion, affecting $\mathbf{q}$. Therefore, differently form \cite{mofid2018adaptive, nicol2011robust}, the cross-coupling term $\mathbf{H}_\sigma( \mathbf q)\ddot{ \mathbf{q}}_{\mathbf u}$ is considered in (\ref{sys_1}), where $\mathbf{H}_\sigma( \mathbf q)$ captures the effects of variations in payload and attitude (cf. Appendix).
\end{remark}
\textbf{Control Problem:} Under Properties 1 and 2, to design a switched adaptive control framework for a vertical operation of a quadrotor following the dynamics (\ref{sys_1}), without any knowledge of system parameters (in line with Remark 1). 

The following section solves this control problem. 

\section{Switched Controller Design}
\begin{assum}[\cite{roy2017adaptive, spong1990adaptive}]\label{assm2}
	The desired trajectories satisfy $\mathbf{q}^d,\dot{\mathbf q}^d, \ddot{\mathbf q}^d \in \mathcal{L}_{\infty}$. Furthermore, $\mathbf{q}, \dot{\mathbf{q}},\ddot{\mathbf q}$ are available for feedback.
\end{assum}
For control design purposes, the dynamics (\ref{sys_1}) is re-arranged as
\begin{equation}\label{sys_2}
\mathbf D_\sigma \ddot{ \mathbf q}+\mathbf E_{\sigma}(\mathbf q,\dot{\mathbf q}, \ddot{\mathbf q},\ddot{\mathbf q}_{\mathbf u},t)=\boldsymbol \tau_{\sigma},
\end{equation} 
which has been obtained by adding and subtracting $\mathbf{D}_\sigma \ddot{\mathbf q}$ to (\ref{sys_1}), where $\mathbf D_\sigma$ is a user-defined constant positive definite matrix and $\mathbf E_{\sigma} \triangleq (\mathbf M_{\sigma}-\mathbf D_\sigma)\ddot{ \mathbf q}+\mathbf C_{\sigma}\dot{\mathbf q}+\mathbf G_{\sigma}+ \mathbf{H}_\sigma\ddot{ \mathbf{q}}_{\mathbf u}+\mathbf{ d}_{\sigma}$. The selection of $\mathbf D_\sigma$ would be discussed later (cf. Remark \ref{select_d}). 


Let $\mathbf {e}(t) \triangleq \mathbf q(t)-\mathbf {q}^d(t)$ be the tracking error, $\boldsymbol \xi (t) \triangleq [\mathbf e(t),~\dot{\mathbf {e}}(t)]$ and $\mathbf r_{\sigma}$ be the filtered tracking error variable defined as
\begin{align}
\mathbf{r}_{\sigma} \triangleq \mathbf{B}^T \mathbf{P}_{\sigma} \boldsymbol \xi, \qquad \sigma \in \Omega \label{r}
\end{align}
where $\mathbf{P}_{\sigma}  >\mathbf 0$ is the solution to the Lyapunov equation $\mathbf{A}_{\sigma}^T \mathbf P_{\sigma} + \mathbf {P_{\sigma} A_{\sigma}}= -\mathbf Q_{\sigma}$ for some $\mathbf{Q}_{\sigma} >\mathbf 0$, $\mathbf A_{\sigma} \triangleq \begin{bmatrix}
\mathbf 0 & \mathbf I\\ 
- \mathbf K_{1\sigma}&  - \mathbf K_{2\sigma}
\end{bmatrix}$ and $\mathbf B \triangleq \begin{bmatrix}
\mathbf 0 &
\mathbf I
\end{bmatrix}^T$. Here, $\mathbf K_{1\sigma}$ and $\mathbf K_{2\sigma}$, are two user-defined positive definite gain matrices and their positive definiteness guarantees that $\mathbf{A}_{\sigma}$ is Hurwitz.

The control law is designed as
\begin{subequations}\label{split_input}
	\begin{align} 
	\boldsymbol \tau_{\sigma} & = \mathbf D_\sigma (-\boldsymbol{\Lambda}_{\sigma} \boldsymbol{ \xi} -\Delta \boldsymbol \tau_{\sigma}+ \ddot{\mathbf{q}}^d),   \label{input}\\
	~ \Delta \boldsymbol \tau_{\sigma}&=\begin{cases}
	{\rho_\sigma}\frac{\mathbf r_\sigma}{|| \mathbf r_\sigma||}       & ~ \text{if } || \mathbf r_\sigma|| \geq \varpi\\
	{\rho_\sigma}\frac{\mathbf r_\sigma}{\varpi}        & ~ \text{if } || \mathbf r_\sigma || < \varpi\\
	\end{cases}, \label{rob}
	\end{align}
\end{subequations}
where $\boldsymbol \Lambda_{\sigma} \triangleq [\mathbf K_{1\sigma}~\mathbf K_{2\sigma}] $; $\varpi >0$ is a user-defined scalar to avoid chattering and the 
design of $\rho_\sigma$ will be discussed later.
Substituting (\ref{input}) in (\ref{sys_2}) yields
\begin{align}
\ddot{\mathbf e}  = -\boldsymbol{\Lambda}_{\sigma} \boldsymbol{ \xi} - \Delta \boldsymbol \tau_{\sigma}+\boldsymbol \chi_{\sigma}, \label{e_ddot}
\end{align}
where $\boldsymbol \chi_{\sigma} \triangleq -\mathbf D_\sigma^{-1} \mathbf{E}_{\sigma}$ is defined as the \emph{overall uncertainty}. Using Properties 1, 2 and Assumption 1 one can verify that 
\begin{equation}
|| \boldsymbol \chi_{\sigma}||\leq \theta_{0 \sigma}^{*} + \theta_{1 \sigma}^{*}  ||  \boldsymbol{ \xi} ||+ \theta_{2\sigma}^{*}  ||  \boldsymbol{ \xi} ||^2+ \theta_{3\sigma}^{*}  ||  \ddot{\overline{\mathbf{ q}}} ||\triangleq \mathbf{Y}_{\sigma}^T \boldsymbol \Theta_{\sigma}^{*} , \label{psi_bound} 
\end{equation}
where $\overline{\mathbf{q}}=[\mathbf{q}^T~\mathbf{q}_{\mathbf u}^T]^T$; $\mathbf{Y}_{\sigma}=[1~||  \boldsymbol{ \xi} ||~ ||  \boldsymbol{ \xi} ||^2~||\ddot{\overline{\mathbf q}}||]^T $ and $\boldsymbol \Theta_{\sigma}^{*}=[\theta_{0\sigma}^{*} ~\theta_{1\sigma}^{*} ~\theta_{2\sigma}^{*}~ \theta_{3 \sigma}^{*}]^T$. The scalars $\theta_{i \sigma}^{*} \in \mathbb{R}^{+}$ $i=0,\cdots,3$ are unknown $\forall \sigma \in \Omega$ and they are defined as 
\begin{align*}
\theta_{0 \sigma}^{*} &= || \mathbf D_\sigma^{-1} || (\overline{g}_\sigma+\overline{d}_\sigma + \overline{c}_\sigma || \dot{\mathbf{q}}^d ||^2), \\
\theta_{1 \sigma}^{*} &= 2  \overline{c}_\sigma || \mathbf D_\sigma^{-1} |||| \dot{\mathbf{q}}^d ||,\\
\theta_{2 \sigma}^{*} &= \overline{c}_\sigma || \mathbf D_\sigma^{-1} ||,  \\
\theta_{3 \sigma}^{*} &= || \mathbf D_\sigma^{-1} \mathbf{M}_\sigma - \mathbf{I} || +  || \mathbf D_\sigma^{-1} || \overline{h}.
\end{align*}
The upper bound structure (\ref{psi_bound}) is obtained by using the following relations $\mathbf{q}=\mathbf{e}+\mathbf{q}^d$, $\dot{\mathbf{q}}=\dot{\mathbf{e}}+\dot{\mathbf{q}}^d$, $||\boldsymbol \xi || \geq || \mathbf{e} ||$ and $||\boldsymbol \xi || \geq || \dot{\mathbf{e}} ||$ (cf. \cite{roy2019adaptive,roy2020adaptive} for details). 

Based on the upper bound structure in (\ref{psi_bound}), the gain $\rho_{\sigma}$ in (\ref{input}) is designed as
\begin{align}
\rho_{\sigma} &= \hat{\theta}_{0 \sigma}+ \hat{\theta}_{1 \sigma} ||  \boldsymbol{ \xi} ||+ \hat{\theta}_{2\sigma}||  \boldsymbol{ \xi} ||^2 +\hat{\theta}_{3 \sigma} || \ddot{\overline{\mathbf{q}}}|| +\zeta_\sigma+ \gamma_\sigma \nonumber\\
&\triangleq \mathbf{Y}^T_{\sigma}\hat{\boldsymbol \Theta}_{\sigma}+ \zeta_\sigma+ \gamma_\sigma, \label{rho}
\end{align}
where $\hat{\boldsymbol \Theta}_{\sigma} \triangleq [\hat{\theta}_{0\sigma} ~ \hat{\theta}_{1\sigma} ~\hat{\theta}_{2\sigma}~ \hat{\theta}_{3 \sigma}]^T$ and $\zeta_\sigma, \gamma_\sigma$ are auxiliary gains needed for closed-loop stabilization. 
Defining $\varrho_\sigma \triangleq ({\lambda_{\min}( \mathbf{Q}_{\sigma} )}/{\lambda_{\max}( \mathbf{P}_{\sigma} )})$, the gains $\hat{\theta}_{i\sigma}, \zeta_\sigma$ and $ \gamma_{\sigma}$ are adapted using the following laws:
\begin{subequations}\label{split_adap}
	\begin{align}
	& \dot{\hat{\theta}}_{jp} =|| \mathbf{r}_p|| || \boldsymbol \xi ||^{j} - \alpha_{jp}{\hat{\theta}}_{j p},~\dot{\hat{\theta}}_{j \overline p}=0,~j=0,1,2 \label{hat_theta1}\\
	&\dot{\hat{\theta}}_{3p} =|| \mathbf{r}_p|| || \ddot{\overline{\mathbf q}} || - \alpha_{3p }{\hat{\theta}}_{3 p},~\dot{\hat{\theta}}_{3 \overline p}=0 \label{hat_theta} \\
	& \dot{\zeta}_{p} = - \left(1 + \hat{\theta}_{3p} || \ddot{\overline{\mathbf q}} || | \mathbf{r}_p || \right) {\zeta}_{ p}+  \bar{\epsilon}_{ p},~ \dot{\zeta}_{\overline p} = 0 \label{zeta}, \\
	& \dot{\gamma}_{p} = 0,~ \dot{\gamma}_{\overline p} = - \left(1 + \frac{\varrho_{\overline{p}}}{2} \sum_{i=0}^{3}{\hat{\theta}}_{i \overline p}^2   \right) {\gamma}_{\overline p} +  {\epsilon}_{\overline p} \label{gamma}, \\
	&\text{with}~\alpha_{i\sigma}> \varrho_{\sigma}/2, ~i=0,1,2,3, \label{alpha}\\
	& \hat{\theta}_{i\sigma} (t_0)>0,~\zeta_{\sigma}(t_0)=\bar{\zeta}_{\sigma} > {\bar \epsilon}_{\sigma }, ~ \gamma_{\sigma}(t_0)=\bar{\gamma}_{\sigma} > {\epsilon}_{\sigma },\label{init} 
	\end{align}
\end{subequations}
where $p$ and $\overline{p} \in \Omega \backslash \lbrace p \rbrace$ denote the active and inactive subsystems respectively; $\alpha_{ip},\bar{\epsilon}_p,\epsilon_{\overline{p}} \in \mathbb{R}^{+}$ are static design scalars and $t_0$ is the initial time. 

We define $\varrho_{M \sigma} \triangleq \lambda_{\max}( \mathbf{P}_{\sigma} ),~ \varrho_{m\sigma} \triangleq \lambda_{\min}( \mathbf{P}_{\sigma} )$, ${\bar{\varrho}_M} \triangleq \max_{\sigma \in \Omega}(\varrho_{M\sigma})$ and ${\underline{\varrho}_m} \triangleq \min_{\sigma \in \Omega} (\varrho_{m\sigma})$. Following Definition 1 of ADT \cite{hespanha1999stability}, the switching law is proposed as
\begin{align}
\vartheta > \vartheta^{*} =  \ln \mu/{\kappa},  \label{sw_law}
\end{align}
where $\mu \triangleq {\bar{\varrho}_M} / \underline{\varrho}_{m}$; $\kappa$ is a scalar defined as $0< \kappa < \varrho$ where $\varrho \triangleq \min_{\sigma \in \Omega}({\lambda_{\min}( \mathbf{Q}_{\sigma} )}/{\lambda_{\max}( \mathbf{P}_{\sigma} )})$.

\begin{remark}[On the use of acceleration measurements]
	Being primarily intended for outdoor applications, use of inertial navigation
	systems with accelerometers are quite common for quadrotor systems \cite{ha2014passivity, lee2012experimental}. 
	Intial pioneering adaptive control designs for (non-switched) systems made use of acceleration measurements \cite{spong1990adaptive}, which was later avoided due to want of sensors. Nevertheless, with the technological advancements and with cheaper prices, usage of acceleration feedback can be found nowadays in many (non-switched) robust and adaptive designs (cf. \cite{roy2019, roy2017adaptive, roy2020adaptive, cst_new} and references therein). 
\end{remark}
\section{Stability Analysis of The Proposed Controller}
\begin{theorem}
	Under Assumption 1 and Properties 1-2, the closed-loop trajectories of system (\ref{e_ddot}) employing the control laws (\ref{split_input}) and (\ref{rho}) with adaptive law (\ref{split_adap}) and switching law (\ref{sw_law}) are Uniformly Ultimately Bounded (UUB). An ultimate bound $b$ on the tracking error $\boldsymbol \xi$ can be found as
	\begin{align}
	b = \sqrt{\frac{{2\bar{\varrho}_M^{(N_0+1)}} \left( \delta + \varpi \delta_1 \right ) }{ {{\underline{\varrho}_m^{(N_0+2)}\left( \varrho - \kappa \right)}} }}, \label{bound} 
	\end{align}
	where the scalars $\delta$ and $\delta_1$ are defined during the proof.
\end{theorem}
\begin{proof}
From (\ref{hat_theta1})-(\ref{gamma}) and the initial conditions (\ref{init}), it can be verified that 
$\exists \underline{\zeta}_{\sigma}, \underline{\gamma}_{\sigma} \in \mathbb{R}^{+}$ such that
\begin{align}
&\hat{\theta}_{i\sigma} (t) \geq 0,~ 0<\underline{\zeta}_{\sigma} \leq \zeta_{ \sigma}(t) \leq \bar{\zeta}_{\sigma},\nonumber\\
\text{and}~&0<\underline{\gamma}_{\sigma} \leq \gamma_{ \sigma}(t) \leq \bar{\gamma}_{\sigma}~~\forall t\geq t_0. \label{low_bound}
\end{align}
	Stability analysis is carried out based on the multiple Lyapunov candidate:
	\begin{align}
	V& = \frac{1}{2} \boldsymbol{\xi}^T \mathbf{P}_{\sigma} \boldsymbol{\xi}  + \sum_{s=1}^{N} \sum_{i=0}^{3} \frac{(\hat{\theta}_{is} -{\theta}_{is}^{*})^2}{2} + \frac{\gamma_{s}}{\underline{\gamma}} + \frac{\zeta_{s}}{\underline{\zeta}} , \label{lyap}
	\end{align}
	where $\underline{\gamma}=\min_{s \in \Omega}( \underline{\gamma}_s )$ and $ \underline{\zeta}=\min_{s \in \Omega}( \underline{\zeta}_s )$. Observing that $\boldsymbol{\Lambda}_{\sigma} \boldsymbol{ \xi} =  \mathbf K_{1\sigma} \mathbf{e} + \mathbf K_{2\sigma} \dot{\mathbf e}$, the error dynamics obtained in (\ref{e_ddot}) becomes
	\begin{align}
	\dot{\boldsymbol \xi} = \mathbf{A}_{\sigma} \boldsymbol \xi + \mathbf{B} \left(\boldsymbol \chi_{\sigma} - \Delta \boldsymbol \tau_{\sigma}\right).  \label{error_dyn}
	\end{align}
	Note that $V(t)$ might be discontinuous at the switching instants and only remains continuous during the time interval between two consecutive switchings. Without loss of generality, the behaviour of $V$ is studied at the switching instant $t_{l+1},~ l \in \mathbb{N}^{+}$. Let an active subsystem be ${\sigma({t_{l+1}^{-}})}$ when $t \in [t_l~~t_{l+1})$ and ${\sigma({t_{l+1}})}$ when $t \in [t_{l+1}~~t_{l+2})$. We have before and after switching
	\begin{align*}
	V({t_{l+1}^{-}}) &= \frac{1}{2}\boldsymbol{\xi}^T({t_{l+1}^{-}}) \mathbf{P}_{\sigma({t_{l+1}^{-}})} \boldsymbol{\xi}({t_{l+1}^{-}})+  \sum_{s=1}^{N} \sum_{i=0}^{3} \left \lbrace \frac{(\hat{\theta}_{is}({t_{l+1}^{-}}) -{\theta}_{is}^{*})^2}{2} + \frac{ \gamma_{s}({t_{l+1}^{-}})}{ \underline{\gamma}}+ \frac{\zeta_{s}({t_{l+1}^{-}})}{ \underline{\zeta} } \right \rbrace ,\\
	V({t_{l+1}}) &= \frac{1}{2}  \boldsymbol{\xi}^T({t_{l+1}}) \mathbf{P}_{\sigma({t_{l+1}})} \boldsymbol{\xi}({t_{l+1}}) +  \sum_{s=1}^{N} \sum_{i=0}^{3} \left \lbrace \frac{ (\hat{\theta}_{is}({t_{l+1}}) -{\theta}_{is}^{*})^2}{2} + \frac{ \gamma_{s}({t_{l+1}})}{ \underline{\gamma}}+ \frac{\zeta_{s}({t_{l+1}})}{ \underline{\zeta}} \right \rbrace,
	\end{align*}
	respectively. Thanks to the continuity of the tracking error $\boldsymbol \xi$ in (\ref{error_dyn}) and of the gains $\hat{\theta}_{i\sigma},\zeta_\sigma$ and $\gamma_{\sigma}$ in (\ref{split_adap}), we have $\boldsymbol \xi({t_{l+1}^{-}}) = \boldsymbol \xi ({t_{l+1}})$, $(\hat{\theta}_{is}({t_{l+1}^{-}}) -{\theta}_{is}^{*})=(\hat{\theta}_{is}({t_{l+1}}) -{\theta}_{is}^{*})$, $\gamma_{s}({t_{l+1}^{-}}) = \gamma_{s}({t_{l+1}})$ and $\zeta_{s}({t_{l+1}^{-}}) = \zeta_{s}({t_{l+1}})$. Further, owing to the facts $ \boldsymbol{\xi}^T({t}) \mathbf{P}_{\sigma(t)} \boldsymbol{\xi}({t}) \leq {\bar{\varrho}_M} \boldsymbol{\xi}^T({t}) \boldsymbol{\xi}({t})$ and $ \boldsymbol{\xi}^T({t}) \mathbf{P}_{\sigma(t)} \boldsymbol{\xi}({t}) \geq {\underline{\varrho}_m}  \boldsymbol{\xi}^T({t}) \boldsymbol{\xi}({t})$, one has 
	\begin{align}
	 V({t_{l+1}}) - V({t_{l+1}^{-}})  &=  \frac{1}{2} \boldsymbol{\xi}^T({t_{l+1}}) ( \mathbf{P}_{\sigma({t_{l+1}})} -  \mathbf{P}_{\sigma({t_{l+1}^{-}})} ) \boldsymbol{\xi}({t_{l+1}}) \nonumber\\
	&\leq  \frac{{\bar{\varrho}_M} - {\underline{\varrho}_m} }{{2\underline{\varrho}_m} }  \boldsymbol{\xi}^T({t_{l+1}}) \mathbf{P}_{\sigma({t_{l+1}^{-}})} \boldsymbol{\xi}({t_{l+1}})  \nonumber\\
	&\leq \frac{{\bar{\varrho}_M} - {\underline{\varrho}_m} }{{\underline{\varrho}_m} } V(t_{l+1}^{-}) \nonumber \\
 \Rightarrow V({t_{l+1}})  &\leq \mu V(t_{l+1}^{-}), \label{mu}
	\end{align}
	with $\mu = {\bar{\varrho}_M} / {\underline{\varrho}_m} \geq 1$. At this point, the behaviour of $V(t)$ between two consecutive switching instants, i.e., when $t \in [t_l~~t_{l+1})$ can be studied. 
	
	We shall proceed the stability analysis for the two cases (i) $|| \mathbf{r}_\sigma || \geq \varpi$ and (ii) $|| \mathbf{r}_\sigma || < \varpi$ using the Lyapunov function (\ref{lyap}). For convenience of notation, let us denote the active subsystem ${\sigma({t_{l+1}^{-}})}$ with $p$ and any inactive subsystem with $\overline{p}$.
	
	\noindent\textbf{Case (i)} $|| \mathbf{r}_\sigma || \geq \varpi$
	
	Using (\ref{psi_bound}), (\ref{error_dyn}), (\ref{split_adap}) and the Lyapunov equation $\mathbf{A}_p^T \mathbf P_p + \mathbf {P}_p \mathbf A_p= -\mathbf Q_{p}$, the time derivative of (\ref{lyap}) yields
	\begin{align}
	\dot{V}  &= \frac{1}{2} \boldsymbol{\xi}^T(\mathbf{A}_p^T \mathbf P_p+ \mathbf {P}_p \mathbf A_p ) \boldsymbol{\xi}+ \boldsymbol \xi^T \mathbf {P}_p \mathbf B \left(\boldsymbol \chi_p -  {\rho_p}\frac{\mathbf r_p}{|| \mathbf r_p||}  \right) +\sum_{s=1}^{N} \sum_{i=0}^{3}\left \lbrace (\hat{\theta}_{is} -{\theta}_{is}^{*})\dot{\hat{\theta}}_{is}+ \frac{\dot{\gamma}_{s}}{ \underline{\gamma}} +\frac{\dot{\zeta}_{s}}{\underline{\zeta}} \right \rbrace \nonumber \\
	&\leq -\frac{1}{2} \boldsymbol{\xi}^T \mathbf Q_p\boldsymbol{\xi} + || \boldsymbol \chi_{p}  || || \mathbf{r}_p ||  - \rho_p{|| \mathbf r_p||} +\sum_{s=1}^{N} \sum_{i=0}^{3}\left \lbrace (\hat{\theta}_{is} -{\theta}_{is}^{*})\dot{\hat{\theta}}_{is}+ \frac{\dot{\gamma}_{s}}{ \underline{\gamma}} +\frac{\dot{\zeta}_{s}}{\underline{\zeta}}\right \rbrace \\
	& \leq  - \frac{1}{2} \boldsymbol{\xi}^T \mathbf Q_p\boldsymbol{\xi}  - \mathbf{Y}^T_p(\hat{\boldsymbol \Theta}_p-\boldsymbol \Theta_p^{*})  || \mathbf{r}_p || +\sum_{s=1}^{N} \sum_{i=0}^{3}\left \lbrace (\hat{\theta}_{is} -{\theta}_{is}^{*})\dot{\hat{\theta}}_{is}+ \frac{\dot{\gamma}_{s}}{ \underline{\gamma}} +\frac{\dot{\zeta}_{s}}{\underline{\zeta}} \right \rbrace. \label{part 2}
	\end{align}
	Using (\ref{hat_theta1})-(\ref{hat_theta})  we have 
	\begin{align}
	\sum_{i=0}^{3}(\hat{\theta}_{ip}-{\theta}_{ip}^{*})\dot{\hat{\theta}}_{ip} & =\sum_{j=0}^{2} (\hat{\theta}_{jp}-{\theta}_{jp}^{*})(|| \mathbf{r}_p || || \boldsymbol \xi ||^{j} - \alpha_{jp}{\hat{\theta}}_{j p})+ (\hat{\theta}_{3p}-{\theta}_{3p}^{*})(|| \mathbf{r}_p || || \ddot{\overline{\mathbf q}} || - \alpha_{3p}{\hat{\theta}}_{3p}) \nonumber\\
	&  = \mathbf{Y}^T_p(\hat{\boldsymbol \Theta}_p-\boldsymbol \Theta_p^{*})  || \mathbf{r}_p ||   + \sum_{i=0}^{3} \left \lbrace\alpha_{ip}{\hat{\theta}}_{ip}{\theta}_{ip}^{*} -\alpha_{ip}{\hat{\theta}}_{ip}^2 \right \rbrace. \label{part 3}
	\end{align}
	Similarly using the facts $\hat{\theta}_{is} \geq 0$, $0< \underline{\zeta}_{s} \leq \zeta_{s}(t),~0< \underline{\gamma}_{s} \leq \gamma_{s}(t)$ from (\ref{low_bound}) and $\underline{\zeta}=\min_{s \in \Omega} ( \underline{\zeta}_{s} ),~\underline{\gamma}=\min_{s \in \Omega} ( \underline{\gamma}_{s} )$, (\ref{zeta}) and (\ref{gamma}) lead to
	\begin{align}
	\frac{\dot{\zeta}_{ p}}{\underline{\zeta}} &=   - \left(1 + \hat{\theta}_{3p} || \ddot{\overline{\mathbf q}} || | \mathbf{r}_p || \right) \frac{{\zeta}_{p}}{\underline{\zeta} } +  \frac{\bar{\epsilon}_{ p}}{\underline{\zeta} }  \nonumber\\ 
	& \leq   - \hat{\theta}_{3p} || \ddot{\overline{\mathbf q}} || | \mathbf{r}_p ||  +  \frac{\bar{\epsilon}_{ p}}{\underline{\zeta} }, \label{part 4_1}\\
	\frac{\dot{\gamma}_{\overline p}}{\underline{\gamma}} &=   - \left(1 + (\varrho_{\overline{p}}/2) \sum_{i=0}^{3}{\hat{\theta}}_{i \overline p}^2   \right) \frac{{\gamma}_{\overline p}}{\underline{\gamma} } +  \frac{{\epsilon}_{\overline p}}{\underline{\gamma} } \nonumber\\ 
	& \leq   - \frac{\varrho_{\overline{p}}}{2} \sum_{i=0}^{3}{\hat{\theta}}_{i \overline p}^2  +  \frac{{\epsilon}_{\overline p}}{\underline{\gamma} }, \label{part 4} 
	\end{align}
	Substituting (\ref{part 3})-(\ref{part 4}) in (\ref{part 2}) yields
	\begin{align}
	\dot{V}\leq &   - \frac{1}{2} \lambda_{\min}(\mathbf Q_p) || \boldsymbol{\xi} ||^2  +\sum_{i=0}^{3} \left \lbrace\alpha_{ip}{\hat{\theta}}_{ip}{\theta}_{ip}^{*} -\bar{\alpha}_{ip}{\hat{\theta}}_{ip}^2 \right  \rbrace \nonumber\\
	&+\frac{\bar{\epsilon}_{ p}}{\underline{\zeta} }  - \left\lbrace \sum_{\forall \overline{p}\in \Omega \backslash \lbrace p \rbrace} \sum_{i=0}^{3} \varrho_{\overline{p}} {\hat{\theta}}_{i \overline p}^2  -  \frac{{\epsilon}_{\overline p}}{\underline{\gamma} } \right \rbrace. \label{part 5}
	\end{align}
	Since $\hat{\theta}_{is} \geq 0$, $\zeta_{s}(t) \leq \bar{\zeta}_{s}$ and $\gamma_{s}(t) \leq \bar{\gamma}_{s}$ by design (\ref{low_bound}), one obtains
	\begin{align}
	V \leq  \frac{1}{2} \lambda_{\max}( \mathbf{P}_{p} ) ||\boldsymbol{\xi}||^2  + \sum_{s=1}^{N} \sum_{i=0}^{3}\frac{(\hat{\theta}_{is}^2 + {\theta_{is}^{*}}^2)}{2}+ \frac{\bar{\gamma}_{s}}{ \underline{\gamma}}+ \frac{\bar{\zeta}_{s}}{ \underline{\zeta}}. \label{lyap_up_bound}
	\end{align}
	Hence, using (\ref{lyap_up_bound}), the condition (\ref{part 5}) is further simplified to
	\begin{align}
	\dot{V} \leq  &- \varrho V+ \frac{\bar{\epsilon}_{ p}}{\underline{\zeta}}  + \sum_{i=0}^{3} \left \lbrace\alpha_{ip}{\hat{\theta}}_{ip}{\theta}_{ip}^{*} -\bar{\alpha}_{ip}{\hat{\theta}}_{ip}^2 \right \rbrace +  \sum_{s=1}^{N} \sum_{i=0}^{3} \frac{\varrho_s{\theta_{is}^{*}}^2}{2}+ \varrho_s \left( \frac{\bar{\gamma}_{s}}{ \underline{\gamma}}+ \frac{\bar{\zeta}_{s}}{ \underline{\zeta}} \right)+\frac{{\epsilon}_{s}}{\underline{\gamma}} , \label{part 6}
	\end{align}
	where $\varrho=\min_{p\in \Omega} \lbrace \varrho_p \rbrace$; $\bar{\alpha}_{ip} =({\alpha}_{ip}-(\varrho_p/2)) >0$ by design from (\ref{alpha}). Again, the following rearrangement can be made
	\begin{align}
	\alpha_{ip}{\hat{\theta}}_{ip} {\theta}_{ip}^{*} -\bar{\alpha}_{ip} \hat{\theta}_{ip}^2 & = - \bar{\alpha}_{ip} \left(  \hat{\theta}_{ip} - \frac{ \alpha_{ip} {\theta}_{ip}^{*}}{2\bar{\alpha}_{ip}} \right)^2 +  \frac{\left( \alpha_{ip} {\theta}_{ip}^{*}\right)^2}{4\bar{\alpha}_{ip}}. \label{part 7_1}
	\end{align}
	We had defined earlier $0< \kappa < \varrho$. Then, using (\ref{part 7_1}), $ \dot{V}(t) $ from (\ref{part 6}) gets simplified to
	\begin{align}
	&\dot{V}(t) \leq - \kappa V(t) - (\varrho - \kappa)V(t) + \delta, \label{part 7}
	\end{align}
	where $\delta \triangleq \max_{p\in \Omega} \left( \sum_{i=0}^{3}\left ( \alpha_{ip} {\theta}_{ip}^{*}\right)^2/({4\bar{\alpha}_{ip}})+(\bar{\epsilon}_{ p}/\underline{\zeta} ) \right) +  \sum_{s=1}^{N} \sum_{i=0}^{3} (\varrho_s/2){\theta_{is}^{*}}^2+ (\varrho_s \bar{\gamma}_{s})/ \underline{\gamma}+(\varrho_s \bar{\zeta}_{s})/ \underline{\zeta}+({\epsilon}_{s}/\underline{\gamma} )$.
	
	\noindent\textbf{Case (ii)} $|| \mathbf{r}_\sigma || < \varpi$
	
	In this case, the time derivative of (\ref{lyap}) yields
	\begin{align}
	\dot{V} &\leq -\frac{1}{2} \boldsymbol{\xi}^T \mathbf Q_p\boldsymbol{\xi} + || \boldsymbol \chi_{p}  || || \mathbf{r}_p ||  - \rho_p\frac{{|| \mathbf r_p||}^2}{\varpi}  +\sum_{s=1}^{N} \sum_{i=0}^{3}\left \lbrace (\hat{\theta}_{is} -{\theta}_{is}^{*})\dot{\hat{\theta}}_{is}+ \frac{\dot{\gamma}_{s}}{\underline{\gamma}}+\frac{\dot{\zeta}_{s}}{\underline{\zeta}} \right \rbrace \\
	& \leq  - \frac{1}{2} \boldsymbol{\xi}^T \mathbf Q_p\boldsymbol{\xi} + || \boldsymbol \chi_{p}  || || \mathbf{r}_p || +\sum_{s=1}^{N} \sum_{i=0}^{3}\left \lbrace (\hat{\theta}_{is} -{\theta}_{is}^{*})\dot{\hat{\theta}}_{is}+ \frac{\dot{\gamma}_{s}}{\underline{\gamma}}+\frac{\dot{\zeta}_{s}}{\underline{\zeta}}\right \rbrace. \label{part 8}
	\end{align}
	Following similar lines of proof as in Case (i) we have
	\begin{align}
	\dot{V} &\leq - \kappa V - (\varrho - \kappa)V + \delta + (\mathbf{Y}^T_p \hat{\boldsymbol \Theta}_p - \hat{\theta}_{3p} || \ddot{\overline{\mathbf q}} ||) || \mathbf{r}_p || \nonumber \\
	& = - \kappa V - (\varrho - \kappa)V + \delta + \sum_{j=0}^{2} \hat{\theta}_{jp}||\boldsymbol \xi ||^j || \mathbf{r}_p || . \label{part 9}
	\end{align}
	From (\ref{r}) one can verify $||\mathbf{r}|| < \varphi \Rightarrow || \boldsymbol \xi || \in \mathcal{L}_{\infty} $ and consequently, the adaptive law (\ref{hat_theta1}) implies $||\mathbf{r}|| , || \boldsymbol \xi || \in \mathcal{L}_{\infty} \Rightarrow \hat{\theta}_{jp} (t) \in \mathcal{L}_{\infty}$, $j=0,1,2$. Therefore, $\exists \delta_1 \in \mathbb{R}^{+}$ such that $\sum_{j=0}^{2} \hat{\theta}_{jp}||\boldsymbol \xi ||^j \leq \delta_1$ $\forall p \in \Omega$ when $||\mathbf r_{p}|| < \varphi$. Hence, replacing this relation in (\ref{part 9}) yields 
	\begin{align}
	&\dot{V}(t) \leq - \kappa V(t) - (\varrho - \kappa)V(t) + \delta + \varpi \delta_1 . \label{part 10}
	\end{align}
	Therefore, investigating the stability results of Cases (i) and (ii), it can be concluded that $\dot{V}(t) \leq - \varrho V(t) $ when
	\begin{align}
	V(t) \geq \mathcal{B} \triangleq \frac{\delta + \varpi \delta_1}{(\varrho - \kappa)}.
	\end{align}
	In light of this, further analysis is needed to observe the behaviour of $V(t)$ between the two consecutive switching instants, i.e., $t \in [t_{l}~t_{l+1})$, for two possible scenarios: 
	\begin{itemize}
		\item[\textbf{(i)}] when $V(t) \geq \mathcal{B}$, we have $\dot{V}(t) \leq - \varrho V(t) $ implying exponential decrease of $V(t)$;
		\item[\textbf{(ii)}] when $V(t) <\mathcal{B}$, no exponential decrease can be derived.
	\end{itemize}
	Behaviour of $V(t)$ is discussed below individually for these two scenarios.
	
	\textbf{Scenario (i):}
	There exists a time, call it $T_1$, when $V(t)$ enters into the bound $\mathcal{B}$ 
	and $N_{\sigma}(t)$ denotes the number of all switching intervals for $t \in [t_0 ~~t_0+T_1)$. Accordingly, for $t \in [t_0 ~~t_0+T_1)$, using (\ref{mu}) and $N_{\sigma}(t_0,t)$ from Definition 1 we have
	\begin{align}
	V(t)   &\leq \exp \left( - \kappa (t-t_{N_\sigma(t)-1})\right) V(t_{N_\sigma(t)-1}) \nonumber\\
	& \leq \mu\exp \left( - \kappa (t-t_{N_\sigma(t)-1})\right) V(t_{N_\sigma(t)-1}^{-}) \nonumber\\
	& \leq \mu\exp \left( - \kappa (t-t_{N_\sigma(t)-1})\right) \nonumber\\
	&~~~\cdot \mu\exp \left( - \kappa (t_{N_\sigma(t)-1}-t_{N_\sigma(t)-2}) \right)V(t_{N_\sigma(t)-2}^{-}) \nonumber\\
	&\qquad \qquad \qquad \qquad \vdots \nonumber\\
	&\leq \mu\exp \left( - \kappa (t-t_{N_\sigma(t)-1})\right)  \mu\exp \left( - \kappa (t_{N_\sigma(t)-1}-t_{N_\sigma(t)-2}) \right) \nonumber\\
	&~~ ~~\cdots \mu \exp \left( - \kappa (t_1-t_0) \right) V(t_0) \nonumber\\
	&= \mu^{N_{\sigma}(t_0,t)} \exp \left(- \kappa (t-t_0) \right)V(t_0)  \nonumber \\
	& = c \left( \exp \left( -\kappa + ({\ln \mu}/{\vartheta}) \right) \right)V(t_0), \label{part 8}
	\end{align}
	where $c \triangleq \exp \left( N_0 \ln \mu \right) $ is a constant. Substituting the ADT condition $\vartheta > \ln\mu / \varrho $ in (\ref{part 8}) yields $V(t) < c V(t_0)$ for $t \in [t_0 ~~t_0+T_1)$. Moreover, as $V(t_0 +T_1) < \mathcal{B}$, one has $V(t_{N_\sigma(t)+1}) < \mu \mathcal{B}$ from (\ref{mu}) at the next switching instant $t_{N_\sigma(t)+1}$ after $t_0+T_1$. This implies that $V(t)$ may be larger than $\mathcal{B}$ from the instant $t_{N_\sigma(t)+1}$. This necessitates further analysis.
	
	We assume $V(t) \geq \mathcal{B}$ for $t \in [t_{ N_\sigma(t)+1}~~ t_0+T_2)$, where $T_2$ denotes the time before next switching. Let $\bar{{N}}_\sigma (t)$ represents the number of all switching intervals for $t \in [t_{N_\sigma(t)+1}~~ t_0+T_2)$. Then, substituting $V(t_0)$ with $V(t_{N_\sigma (t)+1})$ in (\ref{part 8}) and following the similar procedure for analysis as (\ref{part 8}), we have $V(t) \leq c V(t_{ N_\sigma(t)+1}) < c \mu \mathcal{B}$ for $t \in [t_{ N_\sigma (t)+1}~~ t_0+T_2)$. Since $V(t_0+T_2) < \mathcal{B}$, we have $V(t_{ N_\sigma(t)+\bar{ N}_\sigma(t)+2}) < \mu \mathcal{B}$ at the next switching instant $t_{ N_\sigma(t)+\bar{ N}_\sigma(t)+2}$ after $t_0+T_2$. If we follow similar lines of proof recursively, we can come to the conclusion that $V(t) < c \mu \mathcal{B} $ for $t \in [t_0+T_1~~ \infty )$. This confirms that once $V(t)$ enters the interval $[0,\mathcal{B}]$, it cannot exceed the bound $c\mu \mathcal{B}$ any time later with the ADT switching law (\ref{sw_law}).
	
	\textbf{Scenario (ii):} It can be verified that the same argument below (\ref{part 8}) also holds for Scenario (ii). 
	
	Thus, observing the stability arguments of the Scenarios (i) and (ii), it can be concluded that the closed-loop system remains UUB with the control laws (\ref{split_input}) and (\ref{rho}) with the adaptive law (\ref{split_adap}) and switching law (\ref{sw_law}) implying
	\begin{align}
	V(t) \leq \max \left( c V(t_0), c\mu \mathcal{B} \right), ~~\forall t\geq t_0. \label{ub_1}
	\end{align}
	Again, the definition of the Lyapunov function (\ref{lyap}) yields
	\begin{align}
	V(t) \geq ({1}/{2}) \lambda_{\min} (\mathbf P_{\sigma(t)}) || \boldsymbol \xi || ^2 \geq \left({\underline{\varrho}_m }/{2}\right) || \boldsymbol \xi || ^2. \label{ub_2}
	\end{align}
	Using (\ref{ub_1}) and (\ref{ub_2}) we have
	\begin{align}
	|| \boldsymbol \xi || ^2 \leq \left({2}/{\underline{\varrho}_m }\right) \max \left( c V(t_0), c \mu \mathcal{B} \right), ~~\forall t\geq t_0. \label{ub_3}
	\end{align}
	Therefore, using the expression of $\mathcal{B}$ from (\ref{lyap_up_bound}), an ultimate bound $b$ on the tracking error $\boldsymbol \xi$ can be found as (\ref{bound}).
\end{proof}
Note that in collocated design, it is standard to consider the internal dynamics to be stable/bounded a priori \cite{shkolnik2008high,spong1994partial}. In the following, a few design aspects of the proposed mechanism are highlighted.

\begin{remark}[The role of gain $\mathbf D_\sigma$ and its selection]\label{select_d}
	The control law (\ref{input}) relies on the inverse dynamics-type design: in its absence, $\mathbf D_\sigma$ would have been replaced by some nominal value of mass matrix following the conventional design (cf. \cite{roy2019simultaneous,roy2019overcoming}). However, according to Remark 1, such knowledge is unavailable. Again, avoiding inverse dynamics-type design and involving mass matrix in the Lyapunov function eventually lead to a switching law that relies on bound knowledge of mass matrix \cite{roy2019reduced}. These observations highlight the role of $\mathbf D_\sigma$ to realize a switched controller without any knowledge of mass matrix.
	
	Aside being positive definite, the proposed design does not put any restriction on the choice of $\mathbf{D}_\sigma$. It can be observed from (\ref{e_ddot}) that higher values of $\mathbf{D}_\sigma$ reduces the effect of uncertainty $\boldsymbol \chi_\sigma$ on the error dynamics, albeit at the cost of higher control input (cf. (\ref{input})). Therefore, $\mathbf{D}_\sigma$ needs to be designed according to the application requirements. 
\end{remark}
\section{Simulation Results}
\begin{figure}[!t]
	\centering
	\includegraphics[width=\textwidth]{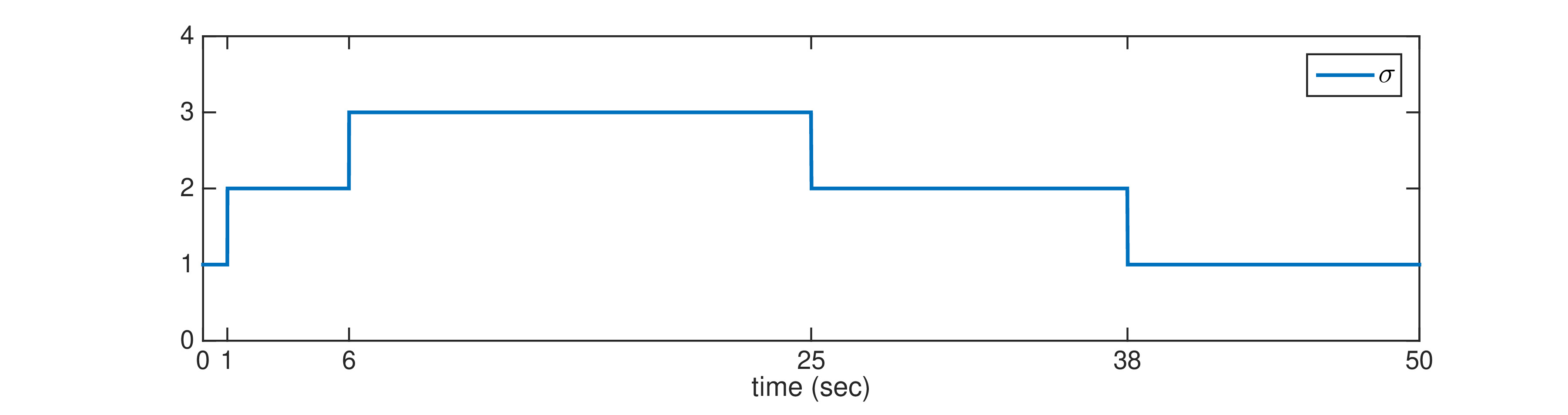}
	\caption{{The switching signal.}}\label{fig:1} 
\end{figure}

\begin{figure}[!t]
	\centering
	\includegraphics[width=\textwidth]{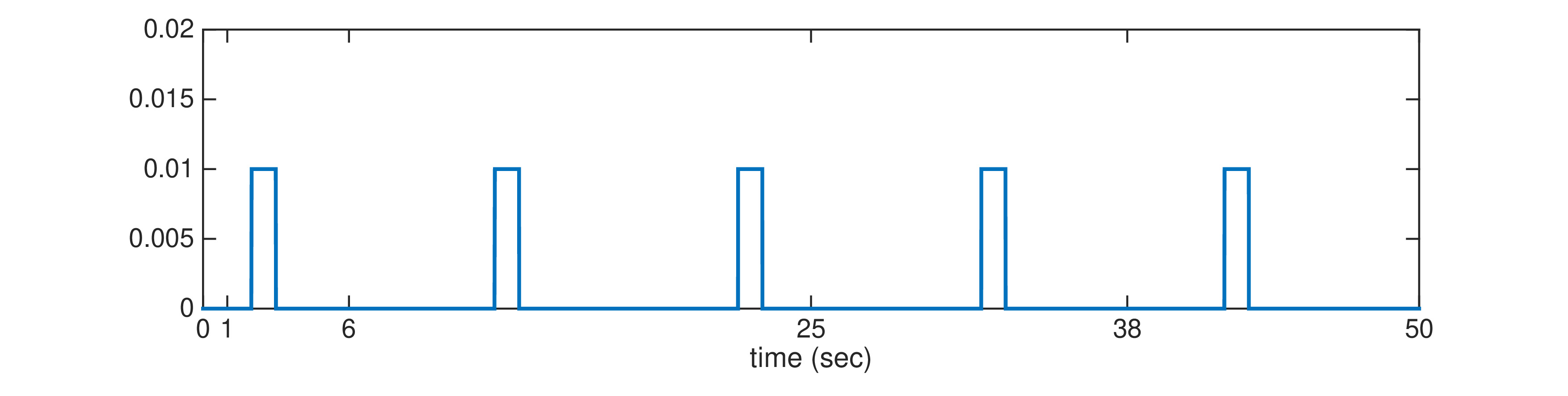}
	\caption{{Pulse type disturbance.}}\label{fig:dist} 
\end{figure}

\begin{figure}[!t]
	\centering
	\includegraphics[width=\textwidth]{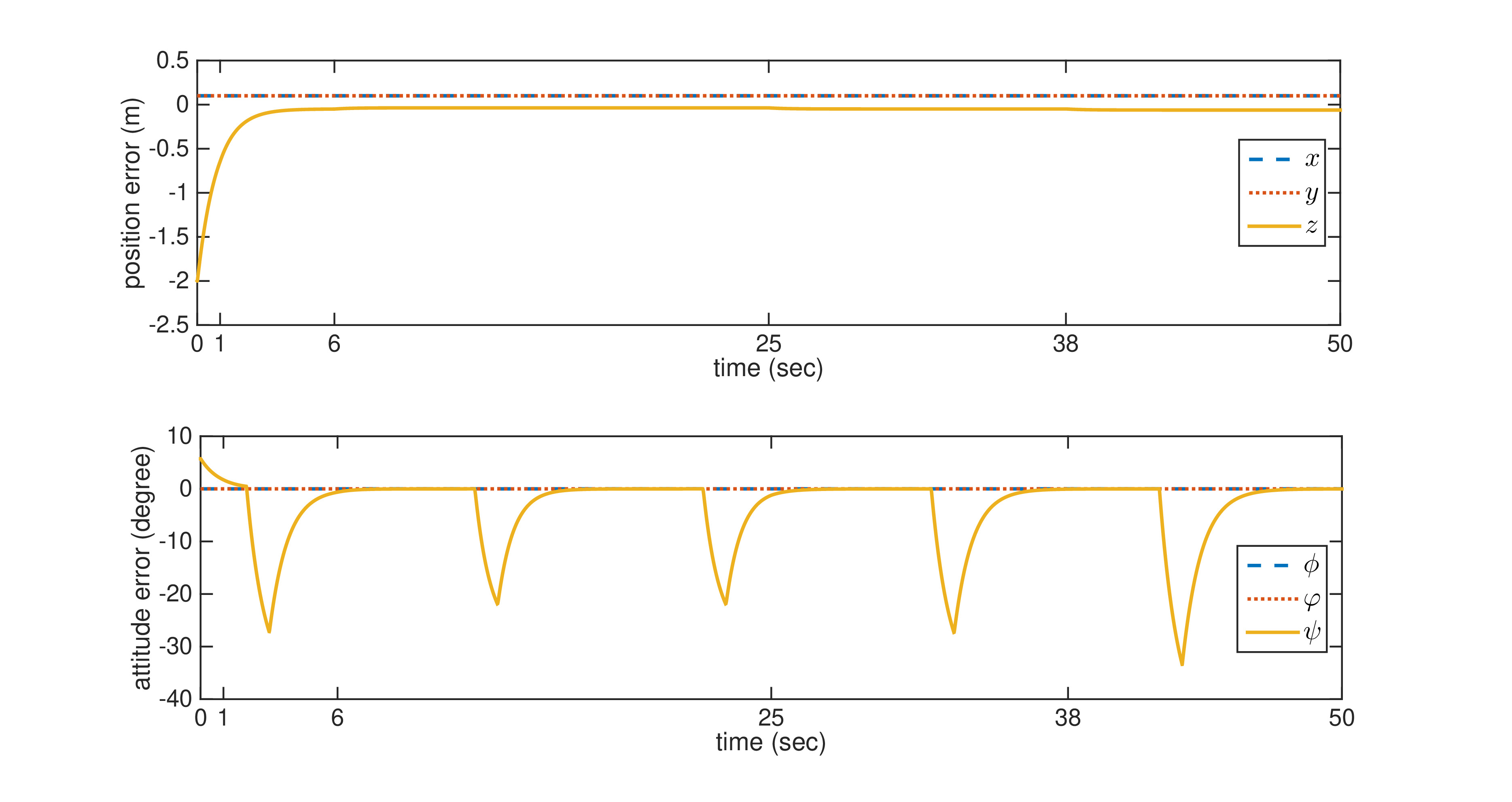}
	\caption{{Tracking performance of the proposed controller.}}\label{fig:2} 
\end{figure}
We test the proposed controller for a switched dynamics scenario as in Fig. \ref{fig:dyn} having three phases (a.k.a subsystems), denoted by $\sigma=1,2,3$ as
\begin{itemize}
\item $\sigma=1$: the quadrotor rests on the ground without any payload;

\item $\sigma=2$: it starts to ascend with an initial payload;

\item $\sigma=3$: a new payload is attached to the quadrotor as it ascends to its desired height.
\end{itemize}

For fruitful verification and as found in many applications (cf. \cite{ye2019observer,roy2018analysis}), we verify the controller on the six degrees-of-freedom quadrotor model. 
The following parametric variations (due to addition of payload) are selected for different subsystems
\begin{align*}
\sigma=1: ~m & =1.5,~I_{xx}=I_{yy}=1.69\cdot10^{-5},~ I_{zz}=3.38\cdot10^{-5},\\
\sigma=2: ~m & =1.6,~I_{xx}=0.011,~ I_{yy}=0.010,~ I_{zz}=1.27\cdot10^{-4},\\
\sigma=3: ~m& =1.7,~I_{xx}=0.032,~ I_{yy}=0.030,~ I_{zz}=2.20\cdot{10}^{-4}.
\end{align*}

The objective for the quadrotor is to lift and drop various payloads at different heights while starting from the ground position. To achieve this task, the following desired trajectories are defined $x^d=y^d=0, z^d= 2+\sin(0.1t), \phi^d=\varphi^d=\psi^d=0$ with an initial position of $ x(0)=y(0)=0.1, z(0)= 0, \phi(0)=\varphi(0),\psi(0)=0.1$. 
Selection of $\mathbf K_{11}=120\mathbf{I}, \mathbf K_{21}=100\mathbf I, \mathbf K_{12}=150\mathbf{I}, \mathbf K_{22}=120\mathbf I,     \mathbf K_{13}=200\mathbf{I}, \mathbf K_{23}=140\mathbf I, \mathbf Q_1=\mathbf Q_2=\mathbf Q_3=2\mathbf I$, $\kappa=0.9\varrho$ yields the ADT $\vartheta^{*}=6.57$sec according to (\ref{sw_law}). Therefore, a switching law $\sigma(t)$ is designed as in Fig. \ref{fig:1} (note that the initial fast switchings are compensated by slower switchings later on). 

Other control parameters are designed as $$\mathbf D_\sigma=\begin{bmatrix}
2 & 0  & 0  & 0 \\ 
0& 1\cdot10^{-4}  & 0 & 0\\ 
0 & 0  & 1\cdot10^{-4}  & 0 \\ 
0 & 0  & 0  & 1\cdot10^{-4} 
\end{bmatrix},$$
$\alpha_{i\sigma}=0.6,\epsilon_{\sigma}=\bar{\epsilon}_{\sigma}=0.005 $ with $i=0,1,2,3$ for all $\sigma$. The initial conditions are selected as $\hat{\theta}_{0\sigma}(0)=1.2,\hat{\theta}_{1\sigma}(0)=1.3,\hat{\theta}_{2\sigma}(0)=1.4,\hat{\theta}_{3\sigma}(0)=1.5 $, $\zeta_\sigma(0)=\gamma_{\sigma}(0)=1$ $\forall \sigma$. The external disturbance is selected as $\mathbf{d}_{\sigma}= \lbrace 0.05\sin(0.5t),0,0,d_p \rbrace$ $\forall \sigma$, where $d_p$ represents a pulse type disturbance (cf. Fig. \ref{fig:dist}) emulating sudden disturbances like gust of wind. It is worth remarking here that, the proposed mechanism being a collocated one, external disturbances should be selected such that the nonactuated dynamics does not become unstable (cf. \cite{shkolnik2008high,spong1994partial}). Therefore, no external disturbances are considered in $(\phi,\varphi)$ as well as in $(x,y)$.

\begin{figure}[!h]
	\centering
	\includegraphics[width=\textwidth]{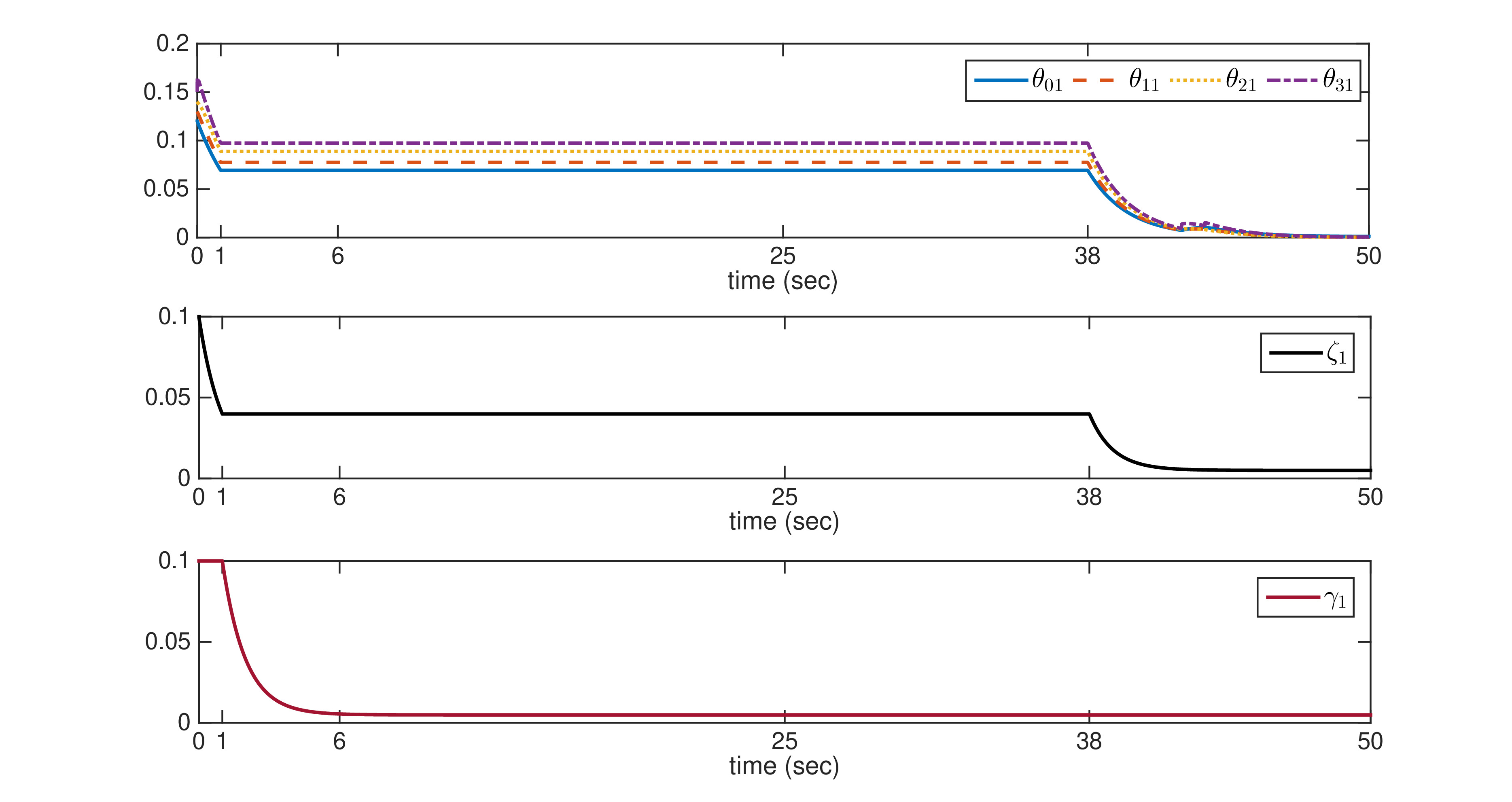}
	\caption{{Gains for subsystem 1.}}\label{fig:3} 
\end{figure}
\begin{figure}[!h]
	\centering
	\includegraphics[width=\textwidth]{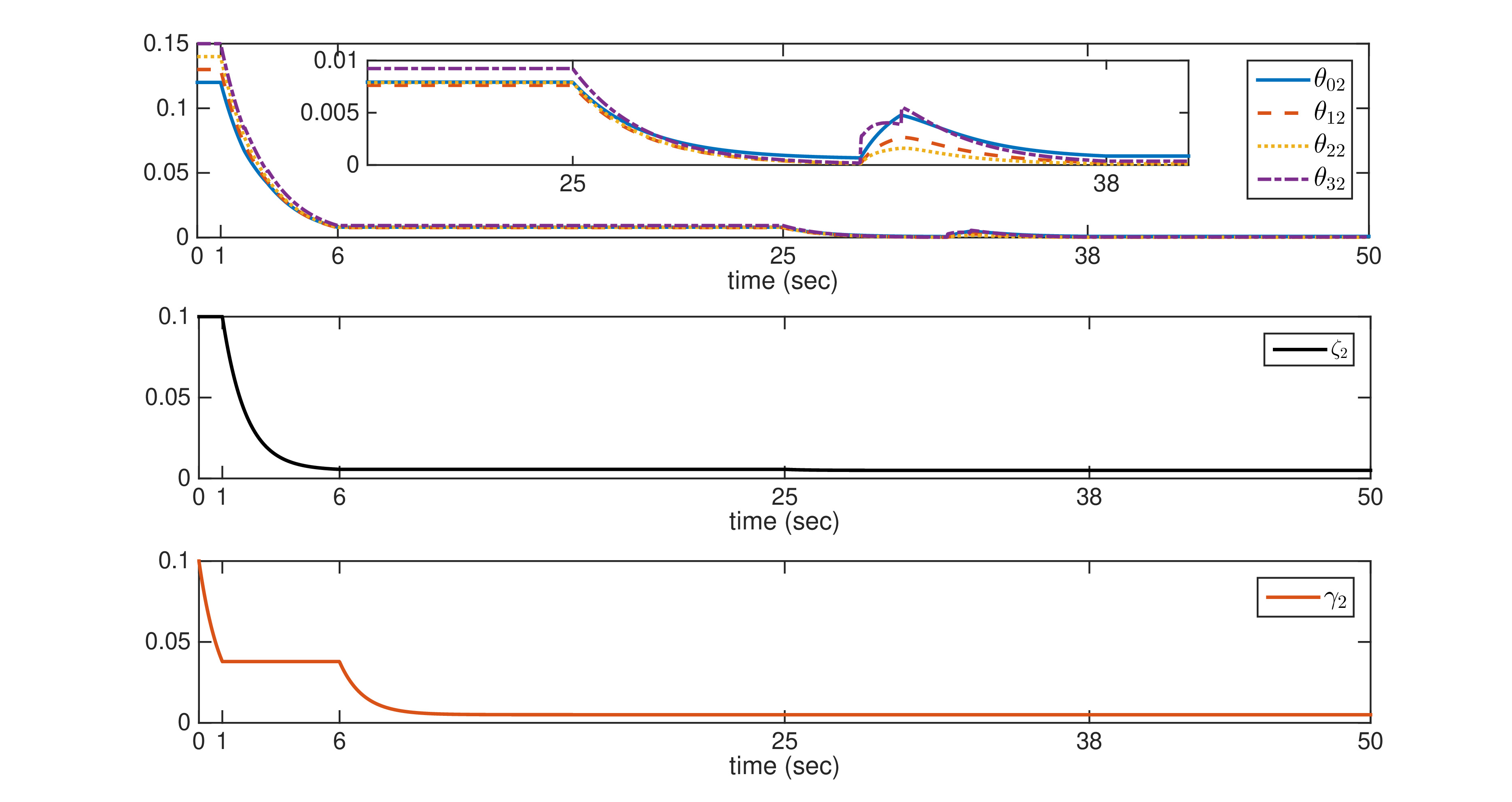}
	\caption{{Gains for subsystem 2.}}\label{fig:4} 
\end{figure}
\begin{figure}[!h]
	\centering
	\includegraphics[width=\textwidth]{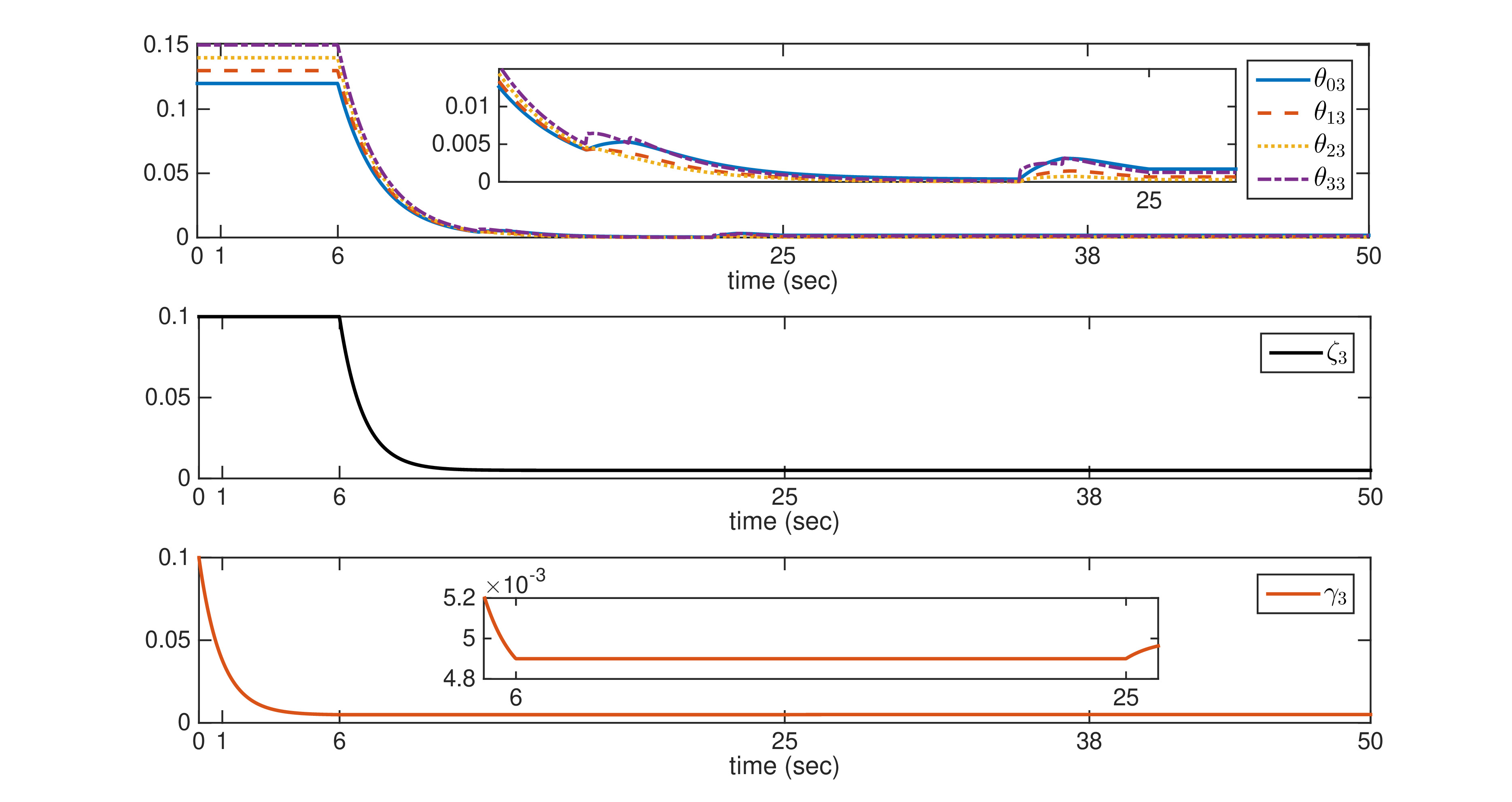}
	\caption{{Gains for subsystem 3.}}\label{fig:5} 
\end{figure}

The performance of the proposed controller is depicted in Fig. \ref{fig:2} in terms of tracking errors, where the attitude errors are reported in degree for better comprehension. In line with the proposed adaptive law (\ref{hat_theta1})-(\ref{gamma}), Figs. \ref{fig:3}-\ref{fig:5} reveal that: (i) for an active subsystem, only the gains $\hat{\theta}_{i {p}}$ and $\zeta_p$ are updated while $\gamma_{p}$ remains constant; (ii) for an inactive subsystem, while $\gamma_{\overline{p}}$ is updated, $\hat{\theta}_{i \overline{p}}$ and $\zeta_{\overline p}$ remain constant. 


\chapter{Conclusions}
\label{ch:conc}
In this thesis, two types of adaptive control was proposed for two operational scenarios of a quadrotor: 
\begin{itemize}
    \item \textbf{Aerial Transportation of Unknown Payloads:} In this scenario, the payload does not change during the flight. Under such a case, an adaptive controller for quadrotors was proposed, which can tackle parametric uncertainties and external disturbances without their a priori knowledge for the application of carrying unknown payloads. The control framework was built to negotiate possibly a priori unbounded state-dependent uncertainties. Closed-loop system stability was established via the notion of uniformly ultimately boundedness. The performance of the proposed controller was verified using Gazebo simulation using RotorS Simulator framework, and comparative simulations confirmed the effectiveness of the proposed scheme against state-of-the-art methods under various scenarios.

\item \textbf{Dynamic Payload Lifting Operations:} Contrary to the first scenario, the payload mass was considered to change for this scenario. Accordingly, a new concept of adaptive control design for quadrotors was introduced for vertical operations, using the framework of average dwell time based switched dynamics. The proposed design did not require any a priori knowledge of the structure and bound of uncertainty. The effectiveness of the concept was validated via simulations with dynamically varying payloads.
\end{itemize}

\section{Future Works}
\begin{itemize}
    \item The current work on dynamic payload lifting operations uses the hovering control dynamics. A more realistic scenario would be to operate in six degrees-of-freedom. In order to solve this problem, a future controller needs to utilise the partly decoupled dynamics for the switched mode control approach.
    
    \item The adaptive controllers developed in this thesis do not tackle the situations of manoeuvring under various motion constraints (e.g., moving through a window, pipeline interior inspection). Therefore, an interesting future work would be to extend the proposed adaptive designs under position and velocity constraints. 
    
    \item Throughout the thesis, the external payloads are considered to be rigidly attached to the quadrotor platform. The case of suspended payload thus needs to be considered in future. 
\end{itemize}
 
\newpage


\label{ch:relatedPubs}
\begin{center}
    
{{\Large{\textbf{Related Publications}}}}

\end{center}
\vspace{1cm}

\section*{{\textbf{\Large{Relevant publications}}}}
\begin{enumerate}
    \item  \textbf{V. N. Sankaranarayanan}, S. Roy and S. Baldi, \textit{``Aerial Transportation of Unknown Payloads: Adaptive Path Tracking for Quadrotors,"} \textbf{2020 IEEE/RSJ International Conference on Intelligent Robots and Systems (IROS),} pp. 7710-7715, 2020. 
    \item  \textbf{V. N. Sankaranarayanan} and S. Roy, \textit{``Introducing switched adaptive control for quadrotors for vertical operations"}, \textbf{Optimal Control Applications and Methods}, vol. 41, no. 6, pp. 1875-1888, 2020. 
\end{enumerate}
\vspace{.3cm}
\section*{{\textbf{\Large{Other publications during MS}}}}
\begin{enumerate}
    \item S. Ganguly, \textbf{*V. N. Sankaranarayanan}, B. V. S. G. Suraj, R. Yadav and S. Roy \textit{``Efficient Manoeuvring of Quadrotor under Constrained Space and Predefined Accuracy,"} \textbf{2021 IEEE/RSJ International Conference on Intelligent Robots and Systems (IROS)}, 2021, accepted for publication. \\
    * \textit{joint first authorship}.
    \item S. Roy, S. Baldi, P. Li and \textbf{V. N. Sankaranarayanan}, \textit{``Artificial-Delay Adaptive Control for Under-actuated Euler-Lagrange Robotics,"} \textbf{IEEE/ASME Transactions on Mechatronics}, 2021.
\end{enumerate}


\bibliographystyle{unsrt}
\bibliography{LaTex/sampleBib} 

\begin{thebibliography}{10}

\bibitem{lee2010geometric}
Taeyoung Lee, Melvin Leok, and N~Harris McClamroch.
\newblock Geometric tracking control of a quadrotor uav on se (3).
\newblock In {\em 49th IEEE Conference on Decision and Control (CDC)}, pages
  5420--5425. IEEE, 2010.

\bibitem{rotors_sim_2016}
Fadri Furrer, Michael Burri, Markus Achtelik, and Roland Siegwart.
\newblock {\em RotorS – A Modular Gazebo MAV Simulator Framework}, volume
  625, pages 595--625.
\newblock 01 2016.

\bibitem{mellinger2014traj}
Kumar~V. Mellinger~D., Michael~N.
\newblock Trajectory generation and control for precise aggressive maneuvers
  with quadrotors.
\newblock In {\em Experimental Robotics. Springer Tracts in Advanced Robotics,
  vol 79}. Springer, Berlin, Heidelberg., 2014.

\bibitem{1302409}
S.~Bouabdallah, P.~Murrieri, and R.~Siegwart.
\newblock Design and control of an indoor micro quadrotor.
\newblock In {\em IEEE International Conference on Robotics and Automation,
  2004. Proceedings. ICRA '04. 2004}, volume~5, pages 4393--4398 Vol.5, 2004.

\bibitem{1389776}
S.~Bouabdallah, A.~Noth, and R.~Siegwart.
\newblock Pid vs lq control techniques applied to an indoor micro quadrotor.
\newblock In {\em 2004 IEEE/RSJ International Conference on Intelligent Robots
  and Systems (IROS) (IEEE Cat. No.04CH37566)}, volume~3, pages 2451--2456
  vol.3, 2004.

\bibitem{boua_pid}
S.~Bouabdallah, P.~Murrieri, and R.~Siegwart.
\newblock Towards autonomous indoor micro vtol.
\newblock In {\em Auton Robot}, volume~18, pages 171--183, 2005.

\bibitem{280180}
Wenwu Yu, ZeFang He, and Long Zhao.
\newblock A simple attitude control of quadrotor helicopter based on
  ziegler-nichols rules for tuning pd parameters.
\newblock {\em The Scientific World Journal}, 2014.

\bibitem{smallanglemodel2}
Long Chen and Gang Wang.
\newblock Attitude stabilization for a quadrotor helicopter using a pd
  controller.
\newblock {\em International Conference on Intelligent Control and Automation
  Science}, 2013.

\bibitem{smallanglemodel3}
S.~Nadda and A.~Swarup.
\newblock Decoupled control design for robust performance of quadrotor.
\newblock {\em Int. J. Dynam. Control 6}, page 1367–1375, 2018.

\bibitem{lee2009asmc}
H.~\& Sastry~S. Lee, D. \& Jin~Kim.
\newblock Feedback linearization vs. adaptive sliding mode control for a
  quadrotor helicopter.
\newblock {\em Int. J. Control Autom. Syst.}, 7:419–428, 2009.

\bibitem{mellinger2011minimum}
Daniel Mellinger and Vijay Kumar.
\newblock Minimum snap trajectory generation and control for quadrotors.
\newblock In {\em 2011 IEEE International Conference on Robotics and
  Automation}, pages 2520--2525. IEEE, 2011.

\bibitem{8688053}
Dawei Shen, Qiang Lu, Min Hu, and Zhiwei Kong.
\newblock Mathematical modeling and control of the quad tilt-rotor uav.
\newblock In {\em 2018 IEEE 8th Annual International Conference on CYBER
  Technology in Automation, Control, and Intelligent Systems (CYBER)}, pages
  1220--1225, 2018.

\bibitem{8815013}
Siddharth Sridhar, Gaurang Gupta, Rumit Kumar, Manish Kumar, and Kelly Cohen.
\newblock Tilt-rotor quadcopter xplored: Hardware based dynamics, smart sliding
  mode controller, attitude hold wind disturbance scenarios.
\newblock In {\em 2019 American Control Conference (ACC)}, pages 2005--2010,
  2019.

\bibitem{kimsurvey}
Jinho Kim, S.~Andrew Gadsden, and Stephen~A. Wilkerson.
\newblock A comprehensive survey of control strategies for autonomous
  quadrotors.
\newblock {\em Canadian Journal of Electrical and Computer Engineering},
  43(1):3--16, 2020.

\bibitem{aerial_mani}
J~Thomas, J~Polin, K~Sreenath, and V.~Kumar.
\newblock Avian-inspired grasping for quadrotor micro uavs.
\newblock {\em ASME 2013 International Design Engineering Technical Conferences
  and Computers and Information in Engineering Conference. Volume 6A: 37th
  Mechanisms and Robotics Conference. Portland, Oregon, USA. August 4–7,
  2013}, 6A, 2013.

\bibitem{tang2015mixed}
Sarah Tang and Vijay Kumar.
\newblock Mixed integer quadratic program trajectory generation for a quadrotor
  with a cable-suspended payload.
\newblock In {\em 2015 IEEE International Conference on Robotics and Automation
  (ICRA)}, pages 2216--2222. IEEE, 2015.

\bibitem{yang2019energy}
Sen Yang and Bin Xian.
\newblock Energy-based nonlinear adaptive control design for the quadrotor uav
  system with a suspended payload.
\newblock {\em IEEE Transactions on Industrial Electronics}, 2019.

\bibitem{fusini2018nonlinear}
Lorenzo Fusini, Thor~I Fossen, and Tor~Arne Johansen.
\newblock Nonlinear observers for gnss-and camera-aided inertial navigation of
  a fixed-wing uav.
\newblock {\em IEEE Transactions on Control Systems Technology},
  26(5):1884--1891, 2018.

\bibitem{kapoutsis2013autonomous}
A~Ch Kapoutsis, Savvas~A Chatzichristofis, Lefteris Doitsidis, J~Borges
  de~Sousa, and Elias~B Kosmatopoulos.
\newblock Autonomous navigation of teams of unmanned aerial or underwater
  vehicles for exploration of unknown static \& dynamic environments.
\newblock In {\em 21st Mediterranean Conference on Control and Automation},
  pages 1181--1188. IEEE, 2013.

\bibitem{nazaruddin2018communication}
Yul~Y Nazaruddin, Augie Widyotriatmo, Tua~A Tamba, Muhammad~S Arifin, and
  Rival~A Santosa.
\newblock Communication-efficient optimal-based control of a quadrotor uav by
  event-triggered mechanism.
\newblock In {\em 2018 5th Asian Conference on Defense Technology (ACDT)},
  pages 96--101. IEEE, 2018.

\bibitem{invernizzi2019dynamic}
Davide Invernizzi, Marco Lovera, and Luca Zaccarian.
\newblock Dynamic attitude planning for trajectory tracking in thrust-vectoring
  uavs.
\newblock {\em IEEE Transactions on Automatic Control}, 65(1):453--460, 2019.

\bibitem{invernizzi2019integral}
Davide Invernizzi, Marco Lovera, and Luca Zaccarian.
\newblock Integral iss-based cascade stabilization for vectored-thrust uavs.
\newblock {\em IEEE Control Systems Letters}, 4(1):43--48, 2019.

\bibitem{sreenath2013trajectory}
Koushil Sreenath, Nathan Michael, and Vijay Kumar.
\newblock Trajectory generation and control of a quadrotor with a
  cable-suspended load-a differentially-flat hybrid system.
\newblock In {\em 2013 IEEE International Conference on Robotics and
  Automation}, pages 4888--4895. IEEE, 2013.

\bibitem{zhao2015nonlinear}
Bo~Zhao, Bin Xian, Yao Zhang, and Xu~Zhang.
\newblock Nonlinear robust sliding mode control of a quadrotor unmanned aerial
  vehicle based on immersion and invariance method.
\newblock {\em International Journal of Robust and Nonlinear Control},
  25(18):3714--3731, 2015.

\bibitem{nersesov2014estimation}
Sergey~G Nersesov, Hashem Ashrafiuon, and Parham Ghorbanian.
\newblock On estimation of the domain of attraction for sliding mode control of
  underactuated nonlinear systems.
\newblock {\em International Journal of Robust and Nonlinear Control},
  24(5):811--824, 2014.

\bibitem{sankaranarayanan2009control}
Velupillai Sankaranarayanan and Arun~D Mahindrakar.
\newblock Control of a class of underactuated mechanical systems using sliding
  modes.
\newblock {\em IEEE Transactions on Robotics}, 25(2):459--467, 2009.

\bibitem{xu2008sliding}
Rong Xu and {\"U}mit {\"O}zg{\"u}ner.
\newblock Sliding mode control of a class of underactuated systems.
\newblock {\em Automatica}, 44(1):233--241, 2008.

\bibitem{Ref:17}
Shafiqul Islam, Peter~X Liu, and Abdulmotaleb El~Saddik.
\newblock Robust control of four-rotor unmanned aerial vehicle with disturbance
  uncertainty.
\newblock {\em IEEE Trans. Ind. Electron.}, 62(3):1563--1571, 2015.

\bibitem{nicol2011robust}
C~Nicol, CJB Macnab, and A~Ramirez-Serrano.
\newblock Robust adaptive control of a quadrotor helicopter.
\newblock {\em Mechatronics}, 21(6):927--938, 2011.

\bibitem{bialy2013lyapunov}
Brendan~J Bialy, Justin Klotz, K~Brink, and Warren~E Dixon.
\newblock Lyapunov-based robust adaptive control of a quadrotor uav in the
  presence of modeling uncertainties.
\newblock In {\em 2013 American Control Conference}, pages 13--18. IEEE, 2013.

\bibitem{dydek2012adaptive}
Zachary~T Dydek, Anuradha~M Annaswamy, and Eugene Lavretsky.
\newblock Adaptive control of quadrotor uavs: A design trade study with flight
  evaluations.
\newblock {\em IEEE Transactions on control systems technology},
  21(4):1400--1406, 2012.

\bibitem{ha2014passivity}
ChangSu Ha, Zhiyuan Zuo, Francis~B Choi, and Dongjun Lee.
\newblock Passivity-based adaptive backstepping control of quadrotor-type uavs.
\newblock {\em Robotics and Autonomous Systems}, 62(9):1305--1315, 2014.

\bibitem{tran2018adaptive}
Trong-Toan Tran, Shuzhi~Sam Ge, and Wei He.
\newblock Adaptive control of a quadrotor aerial vehicle with input constraints
  and uncertain parameters.
\newblock {\em International Journal of Control}, 91(5):1140--1160, 2018.

\bibitem{tian2019adaptive}
Bailing Tian, Jie Cui, Hanchen Lu, Zongyu Zuo, and Qun Zong.
\newblock Adaptive finite-time attitude tracking of quadrotors with experiments
  and comparisons.
\newblock {\em IEEE Transactions on Industrial Electronics}, 2019.

\bibitem{zhao2014nonlinear}
Bo~Zhao, Bin Xian, Yao Zhang, and Xu~Zhang.
\newblock Nonlinear robust adaptive tracking control of a quadrotor uav via
  immersion and invariance methodology.
\newblock {\em IEEE Transactions on Industrial Electronics}, 62(5):2891--2902,
  2014.

\bibitem{roy2019adaptive}
Spandan Roy, Simone Baldi, and Leonid~M Fridman.
\newblock On adaptive sliding mode control without a priori bounded
  uncertainty.
\newblock {\em Automatica}, page 108650, 2019.

\bibitem{roy2020adaptive}
Spandan Roy and Indra~Narayan Kar.
\newblock Adaptive-robust control for systems with state-dependent upper bound
  in uncertainty.
\newblock In {\em Adaptive-Robust Control with Limited Knowledge on Systems
  Dynamics}, pages 117--142. Springer, 2020.

\bibitem{lai2018adaptive}
Guanyu Lai, Zhi Liu, Yun Zhang, CL~Philip Chen, and Shengli Xie.
\newblock Adaptive backstepping-based tracking control of a class of uncertain
  switched nonlinear systems.
\newblock {\em Automatica}, 91:301--310, 2018.

\bibitem{yuan2018robust}
Shuai Yuan, Bart De~Schutter, and Simone Baldi.
\newblock Robust adaptive tracking control of uncertain slowly switched linear
  systems.
\newblock {\em Nonlinear Analysis: Hybrid Systems}, 27:1--12, 2018.

\bibitem{yuan2017adaptive}
Shuai Yuan, Fan Zhang, and Simone Baldi.
\newblock Adaptive tracking of switched nonlinear systems with prescribed
  performance using a reference-dependent reparametrisation approach.
\newblock {\em International Journal of Control}, 92(6):1243--1251, 2019.

\bibitem{lou2018immersion}
Zhi-E Lou and Jun Zhao.
\newblock Immersion-and invariance-based adaptive stabilization of switched
  nonlinear systems.
\newblock {\em International Journal of Robust and Nonlinear Control},
  28(1):197--212, 2018.

\bibitem{chen2018global}
Weisheng Chen, Changyun Wen, and Jian Wu.
\newblock Global exponential/finite-time stability of nonlinear adaptive
  switching systems with applications in controlling systems with unknown
  control direction.
\newblock {\em IEEE Transactions on Automatic Control}, 2018.

\bibitem{ye2021robustifying}
Jun Ye, Spandan Roy, Milinko Godjevac, Vasso Reppa, and Simone Baldi.
\newblock Robustifying dynamic positioning of crane vessels for heavy lifting
  operation.
\newblock {\em IEEE/CAA Journal of Automatica Sinica}, 8(4):753--765, 2021.

\bibitem{sanchez2012continuous}
Anand Sanchez, Vicente Parra-Vega, Chinpei Tang, Fatima Oliva-Palomo, and
  Carlos Izaguirre-Espinosa.
\newblock Continuous reactive-based position-attitude control of quadrotors.
\newblock In {\em 2012 American Control Conference (ACC)}, pages 4643--4648.
  IEEE, 2012.

\bibitem{derafa2012super}
Laloui Derafa, Abdelaziz Benallegue, and L~Fridman.
\newblock Super twisting control algorithm for the attitude tracking of a four
  rotors uav.
\newblock {\em Journal of the Franklin Institute}, 349(2):685--699, 2012.

\bibitem{madani2007sliding}
Tarek Madani and Abdelaziz Benallegue.
\newblock Sliding mode observer and backstepping control for a quadrotor
  unmanned aerial vehicles.
\newblock In {\em 2007 American Control Conference}, pages 5887--5892. IEEE,
  2007.

\bibitem{mofid2018adaptive}
Omid Mofid and Saleh Mobayen.
\newblock Adaptive sliding mode control for finite-time stability of quad-rotor
  uavs with parametric uncertainties.
\newblock {\em ISA transactions}, 72:1--14, 2018.

\bibitem{shtessel2012novel}
Yuri Shtessel, Mohammed Taleb, and Franck Plestan.
\newblock A novel adaptive-gain supertwisting sliding mode controller:
  Methodology and application.
\newblock {\em Automatica}, 48(5):759--769, 2012.

\bibitem{utkin2013adaptive}
Vadim~I Utkin and Alex~S Poznyak.
\newblock Adaptive sliding mode control with application to super-twist
  algorithm: Equivalent control method.
\newblock {\em Automatica}, 49(1):39--47, 2013.

\bibitem{obeid2018barrier}
Hussein Obeid, Leonid~M Fridman, Salah Laghrouche, and Mohamed Harmouche.
\newblock Barrier function-based adaptive sliding mode control.
\newblock {\em Automatica}, 93:540--544, 2018.

\bibitem{roy2020towards}
Spandan Roy and Simone Baldi.
\newblock Towards structure-independent stabilization for uncertain
  underactuated euler--lagrange systems.
\newblock {\em Automatica}, 113:108775, 2020.

\bibitem{fari2019addressing}
Stefano Fari, Ximan Wang, Spandan Roy, and Simone Baldi.
\newblock Addressing unmodelled path-following dynamics via adaptive vector
  field: A uav test case.
\newblock {\em IEEE Transactions on Aerospace and Electronic Systems}, 2019.

\bibitem{yang2019software}
Jun Yang, Ximan Wang, Simone Baldi, Satish Singh, and Stefano Far{\`\i}.
\newblock A software-in-the-loop implementation of adaptive formation control
  for fixed-wing uavs.
\newblock {\em IEEE/CAA Journal of Automatica Sinica}, 6(5):1230--1239, 2019.

\bibitem{roy2019overcoming}
Spandan Roy, Sayan~Basu Roy, Jinoh Lee, and Simone Baldi.
\newblock Overcoming the underestimation and overestimation problems in
  adaptive sliding mode control.
\newblock {\em IEEE/ASME Transactions on Mechatronics}, 2019.

\bibitem{liberzon2003switching}
Daniel Liberzon.
\newblock {\em Switching in systems and control}.
\newblock Springer Science \& Business Media, 2003.

\bibitem{rubisurvey}
Bartomeu Rub{\'\i}, Ramon P{\'e}rez, and Bernardo Morcego.
\newblock A survey of path following control strategies for uavs focused on
  quadrotors.
\newblock {\em Journal of Intelligent \& Robotic Systems}, pages 1--25.

\bibitem{roy2021artificial}
Spandan Roy, Simone Baldi, Peng Li, and Viswa Narayanan.
\newblock Artificial-delay adaptive control for under-actuated euler-lagrange
  robotics.
\newblock {\em IEEE/ASME Transactions on Mechatronics}, 2021.

\bibitem{8362915}
S.~{Yuan}, L.~{Zhang}, O.~{Holub}, and S.~{Baldi}.
\newblock Switched adaptive control of air handling units with discrete and
  saturated actuators.
\newblock {\em IEEE Control Systems Letters}, 2(3):417--422, 2018.

\bibitem{7782779}
S.~{Yuan}, B.~{De Schutter}, and S.~{Baldi}.
\newblock Adaptive asymptotic tracking control of uncertain time-driven
  switched linear systems.
\newblock {\em IEEE Transactions on Automatic Control}, 62(11):5802--5807,
  2017.

\bibitem{roy2019reduced}
Spandan Roy and Simone Baldi.
\newblock On reduced-complexity robust adaptive control of switched
  euler--lagrange systems.
\newblock {\em Nonlinear Analysis: Hybrid Systems}, 34:226--237, 2019.

\bibitem{roy2019simultaneous}
Spandan Roy and Simone Baldi.
\newblock A simultaneous adaptation law for a class of nonlinearly parametrized
  switched systems.
\newblock {\em IEEE Control Systems Letters}, 3(3):487--492, 2019.

\bibitem{roy2020vanishing}
Spandan Roy, Elias~B Kosmatopoulos, and Simone Baldi.
\newblock On vanishing gains in robust adaptation of switched systems: A new
  leakage-based result for a class of {Euler--Lagrange} dynamics.
\newblock {\em Systems \& Control Letters}, 144:104773, 2020.

\bibitem{hespanha1999stability}
Joao~P Hespanha and A~Stephen Morse.
\newblock Stability of switched systems with average dwell-time.
\newblock In {\em Proceedings of the 38th IEEE Conference on Decision and
  Control}, pages 2655--2660, 1999.

\bibitem{spong2008robot}
Mark~W Spong, Seth Hutchinson, and Mathukumalli Vidyasagar.
\newblock {\em Robot dynamics and control}.
\newblock John Wiley \& Sons, 2008.

\bibitem{shkolnik2008high}
Alexander Shkolnik and Russ Tedrake.
\newblock High-dimensional underactuated motion planning via task space
  control.
\newblock In {\em 2008 IEEE/RSJ International Conference on Intelligent Robots
  and Systems}, pages 3762--3768. IEEE, 2008.

\bibitem{spong1994partial}
Mark~W Spong.
\newblock Partial feedback linearization of underactuated mechanical systems.
\newblock In {\em Proceedings of IEEE/RSJ International Conference on
  Intelligent Robots and Systems}, pages 314--321. IEEE, 1994.

\bibitem{roy2017adaptive}
Spandan Roy, Indra~Narayan Kar, Jinoh Lee, and Maolin Jin.
\newblock Adaptive-robust time-delay control for a class of uncertain
  euler--lagrange systems.
\newblock {\em IEEE Transactions on Industrial Electronics}, 64(9):7109--7119,
  2017.

\bibitem{spong1990adaptive}
Mark~W Spong and Romeo Ortega.
\newblock On adaptive inverse dynamics control of rigid robots.
\newblock {\em IEEE Transactions on Automatic Control}, 35(1):92--95, 1990.

\bibitem{lee2012experimental}
Jinoh Lee, Changsun Yoo, Yea-Seok Park, Bumjin Park, Soo-Jin Lee, Dae-Gab
  Gweon, and Pyung-Hun Chang.
\newblock An experimental study on time delay control of actuation system of
  tilt rotor unmanned aerial vehicle.
\newblock {\em Mechatronics}, 22(2):184--194, 2012.

\bibitem{roy2019}
Spandan Roy, Jinoh Lee, and Simone Baldi.
\newblock A new continuous-time stability perspective of time-delay control:
  Introducing a state-dependent upper bound structure.
\newblock {\em IEEE Control Systems Letters}, 3(2):475--480, 2019.

\bibitem{cst_new}
Spandan Roy, Jinoh Lee, and Simone Baldi.
\newblock A new adaptive-robust design for {Time Delay Control} under
  state-dependent stability condition.
\newblock {\em IEEE Transactions on Control Systems Technology}, 2020.

\bibitem{ye2019observer}
Jun Ye, Spandan Roy, Milinko Godjevac, and Simone Baldi.
\newblock Observer-based robust control for dynamic positioning of large-scale
  heavy lift vessels.
\newblock {\em IFAC-PapersOnLine}, 52(3):138--143, 2019.

\bibitem{roy2018analysis}
Spandan Roy, Abhilash Patel, and Indra~Narayan Kar.
\newblock Analysis and design of a wide-area damping controller for inter-area
  oscillation with artificially induced time delay.
\newblock {\em IEEE Transactions on Smart Grid}, 10(4):3654--3663, 2019.

\end{thebibliography}

\end{document}